\documentclass{article}
\usepackage[a4paper, total={6in, 8in}]{geometry}
\usepackage{microtype,graphicx,subfigure}
\usepackage{booktabs, hyperref, url,natbib,caption}
\usepackage{algorithm}
\usepackage[noend]{algorithmic}
\usepackage{amsmath, amsthm, amssymb, mathtools}
\usepackage[capitalize,noabbrev]{cleveref}

\theoremstyle{plain}
\newtheorem{theorem}{Theorem}[section]
\newtheorem{proposition}[theorem]{Proposition}
\newtheorem{lemma}[theorem]{Lemma}

\newtheorem{corollary}[theorem]{Corollary}
\theoremstyle{definition}
\newtheorem{definition}[theorem]{Definition}
\newtheorem{assumption}[theorem]{Assumption}
\theoremstyle{remark}
\newtheorem{remark}[theorem]{Remark}
\usepackage[textsize=tiny]{todonotes}

\theoremstyle{remboldstyle}
\newtheorem*{definition*}{Defintion}
\newtheorem*{theorem*}{Theorem}
\newtheorem*{assu*}{Assumption}
\newtheorem*{assumption*}{Assumption}
\newtheorem*{lemma*}{Lemma}
\newtheorem*{prop*}{Proposition}
\newtheorem*{proposition*}{Proposition}
\newtheorem*{coro*}{Corollary}
\newtheorem{observation}{Observation}

\newcommand{\new}{\color{black}}

\newcommand{\bitem}{\begin{itemize}}
\newcommand{\eitem}{\end{itemize}}
\newcommand{\benum}{\begin{enumerate}}
\newcommand{\eenum}{\end{enumerate}}
\newcommand{\blem}{\begin{lemma}}
\newcommand{\elem}{\begin{lemma}}
\newcommand{\blemn}{\begin{lemma*}}
\newcommand{\elemn}{\begin{lemma*}}

\newcommand{\bdefn}{\begin{definition}}
\newcommand{\bdefnn}{\begin{definition*}}
\newcommand{\edefnn}{\end{definition*}}
\newcommand{\edefn}{\end{definition}}
\newcommand{\bprop}{\begin{proposition}}
\newcommand{\eprop}{\end{proposition}}
\newcommand{\bobsv}{\begin{observation}}
\newcommand{\eobsv}{\end{observation}}

\newcommand{\ps}{\begin{proof}[Sketch]}
\newcommand{\brmk}{\begin{remark}}
\newcommand{\ermk}{\end{remark}}
\newcommand{\bcoro}{\begin{corollary}}
\newcommand{\ecoro}{\end{corollary}}
\newcommand{\bcom}{}

\newcommand{\apx}{approximate}

\newcommand{\arbly}{arbitrarily}
\newcommand{\alg}{algorithm}

\newcommand{\bs}{\backslash}

\newcommand{\cond}{condition}

\newcommand{\distr}{distribution}

\newcommand{\hypo}{hypothesis}

\newcommand{\ho}{\mathbb}

\newcommand{\indep}{independent}

\newcommand{\IOW}{In other words}

\newcommand{\ins}{instance}

\newcommand{\ineq}{inequality}
\newcommand{\ineqs}{inequalities}
\newcommand{\IFT}{It follows that}

\newcommand{\lam}{\lambda}

\newcommand{\ls}{\lesssim}
\newcommand{\lar}{\leftarrow}

\newcommand{\mtg}{martingale}
\newcommand{\mtx}{matrix}

\newcommand{\Omg}{\Omega}

\newcommand{\rb}{\right}

\newcommand{\lb}{\left}

\newcommand{\prb}{probability}
\newcommand{\prbs}{probabilities}

\newcommand{\parti}{particular}

\newcommand{\pmt}{parameter}

\newcommand{\rv}{random variable}

\newcommand{\rar}{\rightarrow}

\newcommand{\real}{\mathbb{R}}
\newcommand{\resp}{respectively}

\newcommand{\sat}{satisfy}
\newcommand{\sats}{satisfies}
\newcommand{\satd}{satisfied}

\newcommand{\sse}{\subseteq}
\newcommand{\sps}{suppose}
\newcommand{\Sps}{Suppose}

\newcommand{\strfwd}{straightforward}

\newcommand{\sys}{system}

\newcommand{\unif}{uniform}

\newcommand{\wh}{\widehat}
\newcommand{\wt}{\widetilde}

\newcommand{\xulie}{sequence}

\let\eps\varepsilon

\begin{document}
\title{Multi-Armed Bandits with Interference}
\author{Su Jia, Peter Frazier, and Nathan Kallus\\
Cornell University}

\date{}

\maketitle
\begin{abstract}
Experimentation with interference poses a significant challenge in contemporary online platforms. 
Prior research on experimentation with interference has concentrated on the final output of a policy. Cumulative performance, while equally crucial, is less well understood.
To address this gap, we introduce the problem of {\em Multi-armed Bandits with Interference} (MABI), where the learner assigns an arm to each of $N$ experimental units over a time horizon of $T$ rounds.
The reward of each unit in each round depends on the treatments of {\em all} units, where the influence of a unit decays in the spatial distance.
Furthermore, we employ a general setup in which the reward functions are chosen by an adversary and may vary arbitrarily between rounds and units.
We first show that switchback policies achieve an optimal {\em expected} regret $\tilde O(\sqrt T)$ against the best fixed-arm policy.
Nonetheless, the regret (as a random variable) for any switchback policy suffers a high variance, as it does not account for $N$. 
We propose a cluster randomization policy whose regret (i) is optimal in {\em expectation} and (ii) admits a high probability bound that vanishes in $N$.
\end{abstract}

\section{Introduction} 
Online controlled experiments (``A/B tests'') have become a standard practice to evaluate the impact of a new product or service change before wide-scale release.
Although A/B tests work well in many situations, they often fail to provide reliable estimates in the presence of {\em interference}, where the treatment of one unit affects the outcome of another.
For example, suppose that a ride-sharing firm develops a new pricing algorithm. 
If the firm randomly assigns half of the drivers to the new algorithm, these drivers will alter their behaviors, which will impact the common pool of passengers and consequently the drivers not assigned the new algorithm.

Most previous work on experimentation with interference focused on the quality of the {\em final} output, e.g., the mean-squared error of the estimator (e.g., \citealt{ugander2013graph}) or the $p$-value in a \hypo\ testing setting (e.g., \citealt{athey2018exact}).
On the other hand, the cumulative performance of the AB testing policies is also considerably important in practice, considering the scale of experiments on modern online platforms. 
However, this perspective is often overlooked.

This motivates us to study the aspect of cumulative reward maximization in A/B testing with interference. 
We employ a batched adversarial bandit framework. 
{\new We are given a set $U\sse \real^2$ of $N$ {\em units} representing, for example, users in an online platform, each with a known, fixed location.}
We are also given a set of $k$ {\em treatment arms} (or {\em arms}) and a time horizon with $T$ rounds.
In each round, the learner assigns one arm to each unit and collects an observable reward. 
The goal is to maximize the total reward.

The reward in each unit in each round is governed by a {\em reward function}, which is secretly chosen by an adversary at the beginning.
To capture interference, the reward is determined by the treatments of {\bf all} units. 
In other words, each reward function is defined on $[k]^U$ rather than on $[k]$.

Similarly to causal inference with network interference, efficient learning is impossible without additional structures. 
In line with reality, we assume that the unit-to-unit interference decreases in their (spatial) distance.
Specifically, we employ \citet{leung2022rate}'s {\em decaying interference property}: The rewards of any two treatment vectors $z,z'\in [k]^U$ for a unit $u\in U$ are close if $z,z'$ are identical in a neighborhood of $u$. 
In addition, the larger this neighborhood, the closer these rewards become.


\subsection{Our Contributions}
We contribute to the literature of multi-armed bandits and causal inference with interference in the following ways.
\benum 
\item {\bf Bridging Online Learning and Causal Inference.} {\new We incorporate interference into  the study of {\em multi-armed bandits} (MAB) by formulating the problem of {\em Multi-Armed Bandits with Interference} (MABI). 
The ordinary MAB problem can therefore be seen as a special case of MABI under SUTVA.
Our formulation is fairly general, imposing no constraints on the non-stationarity or similarity in reward functions between units.
At the heart of our policy is an estimator that generalizes both the one from \citealt{kocak2014efficient} (for adversarial bandits) and the one from \citealt{leung2022rate} (for  inference with spatial interference); we explain this further in Bullet 3.}
\item {\bf Optimal Expected Regret.} We characterize the minimax expected regret as follows.
\benum 
\item {\bf Upper Bound.} We show that any adversarial bandits policy can be converted into a switchback policy (i.e., a policy that assigns all units the same arm in each round) for the MABI problem, and this conversion preserves the regret; see \cref{prop:reduction_to_adv_bandits}.
In \parti, this implies an $\widetilde O(\sqrt {kT})$ upper bound on the expected regret; see \cref{coro:ub_MABI}. 
\item {\bf Lower Bound.} We show that for any (possibly non-switchback) policy, there is a MABI instance on which the policy suffers an $\Omg(\sqrt{kT})$ expected regret; see \cref{thm:lb}. 
This suggests that $N$ does not help reduce the expected regret.
{\new This lower bound relies on a novel combinatorial result (\cref{prop:homotopy}): Given a function $f$ defined at the two corners ${\bf 0}^U, {\bf 1}^U$ of the hypercube $\{0,1\}^U$, we can extend it to the entire hypercube subject to the decaying interference constraint.}
\eenum
\item {\bf High Probability Regret Bound.} 
Although the regret (as a random variable) of a switchback policy may be optimal in expectation, it suffers a high variance. 
To address this, we propose a policy that {\new integrates the idea of (i) implicit exploration from  adversarial bandits and (ii) clustered randomization from causal inference.} It involves the following components.
\benum 
\item {\bf The Robust Randomized Partition.} \citet{leung2022rate} achieved an optimal statistical performance for inference with spatial interference using a uniform spatial clustering, where each arm is assigned to each cluster with probability $p=\Theta(1)$.
Unfortunately, in our problem, $p$ may be exceptionally small due to weight updates, leading to a high variance in the estimator. 
To address this, we introduce the {\em Robust Randomized Partition}. 
{\new The key idea is to ``perturb'' the cluster boundaries {\bf randomly}.}
This increases the minimum exposure probability from $p^{O(1)}$ to $\Omega(p)$, and consequently reduces the variance. 
\item {\bf The HT-IX Estimator.} 
To achieve a high probability regret bound for adversarial bandits, \citet{kocak2014efficient} proposed the EXP3-IX policy which modified the EXP3 policy by adding an {\em implicit exploration} (IX) parameter to the propensity weight. 
On the other hand, to estimate the average treatment effect under spatial interference, \citet{leung2022rate} proposed a truncated Horvitz-Thompson (HT) estimator for the (single-period) inference problem.
We introduce the {\em Horvitz-Thompson-IX} (HT-IX) estimator which unifies these two estimators.
\item {\bf High Probability Regret Bound.} We show that the EXP3 policy based on the HT-IX estimator (i) has an optimal expected regret, and (ii) its tail mass above for any threshold  vanishes as $N\rar \infty$; see \cref{thm:main}. 
In contrast, this result is impossible for any switchback policy: The tail mass of the regret beyond any given threshold does {\bf not} decrease in $N$.
\eenum
\eenum 

Finally, we emphasize that, from a practical perspective, the market size $N$ should be considered substantially greater than the number $T$ of rounds. Therefore, it is vital to leverage $N$ in the design and estimator.
We next review the previous literature on adversarial bandits and causal inference under interference.

\subsection{Related Work}
Experimentation is a widely deployed learning tool in online commerce that is easy to execute \citep{kohavi2017surprising,thomke2020building,larsen2023statistical}.
As a key challenge, the violation of the so-called {\em Stable Unit Treatment Value Assumption} (SUTVA) has been viewed as problematic for online platforms \citep{blake2014marketplace}.
This problem has been extensively studied in statistics (e.g., \citealt{hudgens2008toward,aronow2017estimating,eckles2017design,basse2018analyzing,basse2019randomization,li2022random,hu2022average,leung2023network,hu2022switchback}), operations research (e.g., \citealt{johari2022experimental,bojinov2023design,farias2022markovian,holtz2024reducing,candogan2024correlated,jia2023clustered}), computer science (e.g., \citealt{ugander2013graph,ugander2023randomized,yuan2021causal}) and medical research \citep{tchetgen2012causal}. 
Some recent surveys include \citealt{bajari2023experimental, larsen2023statistical}.

Many works tackle this problem by assuming that interference is summarized by a low-dimensional exposure mapping and that units are individually randomized to treatment or control by Bernoulli or complete randomization \cite{manski2013identification,toulis2013estimation,aronow2017estimating,basse2019randomization,forastiere2021identification}. 
To improve estimator precision, some work departed from unit-level randomization and introduced cluster correlation in treatment assignments. {\new This is usually done by either (i) grouping the units in a network into clusters \citep{ugander2013graph,jagadeesan2020designs,leung2022rate,leung2023network} or (ii) grouping time periods into blocks (``switchback'') \citep{bojinov2023design,hu2022switchback,jia2023clustered}.} 
However, these works usually focus on the quality of the final output, such as the bias and variance of the estimator \citep{ugander2013graph,leung2022rate} and $p$-values for hypotheses testing \cite{athey2018exact}.

While existing literature has primarily focused on the final output, the cumulative performance remains less well understood.
A natural framework is  {\em mult-armed bandits} (MAB) \citep{lai1985asymptotically}.
{\new These work focus on the cumulative performance but often overlooks the interference element}. 
There are three lines of work in MAB that are most related to this work: (i) adversarial bandits, (ii) multiple-play bandits and (iii) combinatorial bandits.

Particularly related is the adversarial bandit problem.
Many policies for adversarial bandits are built on the idea of weight update \citep{vovk1990aggregating,littlestone1994weighted}, first introduced for the full-information setting (i.e., the {\em best expert problem}). 
\citet{auer1995gambling} considered the bandit feedback version and proposed a forced exploration version of the EXP3 policy.
\citet{stoltz2005incomplete} observed that the policy achieves the optimal expected regret even without additional exploration.

High-probability bounds for adversarial bandits were first provided by 
\cite{auer2002nonstochastic} and explored in a more generic way by \cite{abernethy2009beating}.
In \parti, the idea to reduce the variance of importance-weighted estimators has been applied in various forms (e.g., \cite{Ionides2008truncated,bottou2013counterfactual}), and was first introduced to bandits by
\citet{kocak2014efficient}.
Subsequently, \citet{neu2015explore} showed that this algorithm admits high-probability regret bounds.

Another closely related line is 
{\em multiple-play bandits} \citep{anantharam1987asymptotically}, where the learner plays multiple arms per round and observes each of their feedback.
In these work, the number of arms played in each round can be viewed as the ``$N$'' in our problem
\citep{chen2013combinatorial,komiyama2015optimal,lagree2016multiple,jia2023short}. 
Another related line of work is multi-agent RL \citep{kanade2012distributed,busoniu2008comprehensive,zhang2021multi}. Our problem differs from these work in that each ``agent'' behaves completely passively.

Finally, since the reward function is defined on the hypercube $[k]^U$, our work is also related to {\em combinatorial bandits} \citep{cesa2012combinatorial} where the action set is a subset of a binary hypercube.
However, most work in this area considers linear objectives, where our reward functions are only assumed to satisfy the decaying interference assumption, which is much more general than linear functions.
A recent line of work focus on combinatorial bandits with non-linear reward functions. 
However, most of these work either assumes a stochastic setting \citep{agrawal2017thompson,kveton2015combinatorial}, or consider an adversarial setting with a specialized class of reward functions, such as polynomial link functions \citep{han2021adversarial}.

{\new Recently, \cite{agarwal2024multi} considered an online learning problem with the same motivation. 
Similarly to our work, they also incorporated interference by defining the reward function on the hypercube $[k]^U$. 
However, they focused on network interference, where the reward of a unit is determined by the treatment on the immediate neighbors (and itself, of course).
Moreover, their reward function is fixed over time. Finally, our work defines the regret against the best fixed-arm policy where all units and all rounds are assigned the same arm. 
In contrast, they consider the best fixed assignment $z\in [k]^U$ in hindsight.}

\section{Formulation and Assumptions} 
We consider a {\em multiple-play} (i.e., multiple arms are played in each round) adversarial bandits setting.
Consider a set $U$ of {\em units}, $T$ treatment decision {\em rounds} and $k$ {\em treatment arms} (or, simply, {\em arms}). 
For each round $t\in [T]$ and unit $u\in U$ there is a {\em reward function} $Y_{ut}: [k]^U \rar [0,1]$ that can depend on the treatment assignment $z\in [k]^U$ of all units (not just that of $u$).
In each round $t$, the learner selects a {\em treatment assignment} $Z_t\in [k]^U$ and observes a reward of $Y_{ut}(Z_t)$ for each $u\in U$. \footnote{referred to as the {\em bandit feedback} in the literature of MAB.}

\subsection{Policy}
The treatment assignment depends only on the past. 
Formally, a {\em policy} (or, adaptive design) $\pi=(\pi_1,\dots,\pi_T)$ consists of mappings \[\pi_t: ([k]^U\times [0,1]^U)^{t-1}\to\Delta([k]^U).\] 
In round $t$, we draw $Z_t$ from the distribution 
\[\pi_t(Z_1,(Y_{u1}(Z_1))_{u\in U},\dots,Z_{t-1},(Y_{u,t-1}(Z_{t-1}))_{u\in U}).\]
As in adversarial bandits, we aim to control the loss compared to the best fixed arm.
When $N=1$, the following notion of regret can be easily verified to coincide with the regret in adversarial bandits.

\bdefn[Regret] \label{def:regret}
The {\em regret} of a policy $Z$ is defined as \[{\rm Reg}(Z):= \max_{a\in [k]} \lb\{{\rm Reg}(Z, a)\rb\},\] where for each arm $a\in [k]$, we define \[{\rm Reg}(Z, a):= \sum_{t=1}^T \frac 1N\sum_{u\in U} \lb(Y_{ut}\lb(a\cdot {\bf 1}^U\rb) - Y_{ut}(Z_t)\rb).\]
\edefn 

We focus on bounding the regret in expectation or, preferably, in high probability. 
The reward functions are chosen ``secretly'' in advance in the specific sense that our bounds on regret will hold for \textit{any} reward functions (possibly subject to some constraints we discuss next). 
Thus, the bounds hold even for the worst-case reward functions chosen by an adversary.

\subsection{Spatial Interference}
{\new Thus far, the model allows for unrestricted interference in the sense that $Y_{ut}(z)$ may vary  arbitrarily in any coordinate of $z$. 
To obtain a positive result on asymptotic inference, it is necessary to impose restrictions on interference to establish some form of weak dependence across unit outcomes. 
The existing literature focuses on the restrictions captured by $\kappa$-neighborhood exposure mappings, which imply that the arm assigned to $v$ (i.e., $z_v$) can only interfere with $Y_{ut}$ if the distance between $u,v$ is at most $\kappa$, which is somewhat restrictive.

In many applications, the effect of a treatment diffuses primarily through physical interaction, such as a promotion on a ride-sharing platform or a discount on a food delivery platform. 
\citet{leung2022rate} addressed this by proposing a model that allows for interference between any two units, with an intensity that decayss in the distance.
The following assumption is identical to that in their \S 2.1 (up to re-scaling).

\begin{assumption}[Scaling of the Bounding Box, \cite{leung2022rate}]\label{assu:bounding_box} We assume that $U\sse [-b_N, b_N]$ where $b_N =O(\sqrt N)$ and $d (u,v)\ge 1$ for any $u,v\in U$ where $d$ is the sup norm.\footnote{This choice is not essential. Our results hold (up to $O(1)$) for any $L_p$-norm ($p\ge 1$) since $U\sse \real^2$.}
\end{assumption}
}


\citet{leung2022rate} posit that if two assignments $z,z'$ are identical on a ball-neighborhood of $u$, then the rewards of $u$ under $z,z'$ are close.
To formalize, we denote the radius-$r$ (open) {\em ball} as $B(u,r):= \{v\in U\mid d(u,v)<r\}$.
For clarity, $B(u,r)$ does not contain units that are {\bf exactly} distance $r$ away. 
In particular, $B(u,0)=\emptyset$ and $B(u,1)=\{u\}.$

\bdefn[Decaying Interference Property]\label{assu:DIP}
Let $\psi:[0,\infty)\rar [0,\infty)$ be a non-increasing function.
A MABI instance \sats\ the {\em $\psi$-decaying interference property} (or {\em $\psi$-DIP}), if for any $r\ge 0, u\in U, t\in[T]$ and $z,z'\in [k]^U$ with $z_{B(u,r)} = z'_{B(u,r)}$,\footnote{For any subset $S\sse U$, we denote $z_S = (z_u)_{u\in S}$.}
we have \[|Y_{ut}(z)- Y_{ut}(z')| \le \psi(r).\] 
\edefn 

\brmk[SUTVA] Consider $\psi(r)={\bf 1}(r=0)$. Then, taking any $r>0$ in \cref{assu:DIP}, we have $|Y_{ut}(z)- Y_{ut}(z')| \le \psi(1) = 0$ for any $z,z'$ that are identical on $B(u,1) = \{u\}$. 
\IOW, $Y_{ut}(z)$ only depends on $z_u$.
This is the well-known {\em Stable Unit Treatment Values Assumption} (SUTVA) \citep{rubin1978bayesian}. \ermk

\section{Expected Regret}
In this section, we show that the minimax {\em expected} regret for the MABI problem is $\widetilde \Theta (\sqrt T)$. 
We emphasize that this result holds for {\bf all} $N$ (not fixed as constant!), but $N$ does not show up in the regret. This is due to the normalization factor $1/N$ in the definition of regret; see \cref{def:regret}.

\subsection{Upper Bound}\label{subsec:UB}
We begin by observing that the MABI problem is equivalent to adversarial bandits when we restrict ourselves to a special type of policy - the switchback policy - which selects the same arm for all units. {\new These policies are widely applied in practice \citep{doordash,amazon} and have been extensively studied \citep{bojinov2023design,hu2022switchback,xiong2023data}}.

\bdefn[Switchback Policy] A policy $Z=(Z_t)_{t\in [T]}$ is a {\em switchback policy} if there is an adversarial bandits policy $(A_t)$ with $Z_t = A_t \cdot {\bf 1}^U$ a.s. for each $t\in [T]$.
\edefn

By the above definition, any adversarial bandit policy $(A_t)$ induces a switchback policy $(Z_t)$ where $Z_t = A_t \cdot {\bf 1}^U$.
Moreover, this reduction preserves the regret.

\begin{proposition}[Reduction to Adversarial Bandits]
\label{prop:reduction_to_adv_bandits}
Let $A=(A_t)$ be an adversarial bandits policy with regret $r(T)$.
Then, the MABI policy $Z=(Z_t)$ given by $Z_t = A_t \cdot {\bf 1}^U$ satisfies ${\rm Reg}(Z) = r(T).$
\end{proposition}

Note that for adversarial bandits, the EXP3\footnote{means ``EXPlore and EXPloit with EXPonential weights''} policy \cite{auer1995gambling} has a $\widetilde O(\sqrt {kT})$ regret, so: 

\bcoro[Upper Bound on Expected Regret] 
\label{coro:ub_MABI}
Let $Z$ be the switchback policy induced by the EXP3 policy, then ${\rm Reg}(Z) =\widetilde O(\sqrt{kT})$.
\ecoro

\subsection{Lower Bound on the Expected Regret}

{\new The meticulous reader may have noticed that the bound in \cref{coro:ub_MABI} does not involve $N$.
This is actually quite reasonable. In fact, a switchback policy treats the entire \sys\ as a whole, and therefore does not take advantage of the market size $N$.}
Can we improve this bound by leveraging $N$?
Surprisingly, the answer is no.

\begin{theorem}[Lower Bound on the Expected Regret]\label{thm:lb}
Fix any non-increasing function $\psi$ with $\psi(0)=1$ and $\lim_{x\rar \infty}\psi(x)= 0$. 
Then for any MABI policy $Z$, there exists a MABI \ins\ $\cal I$ \sat ing the $\psi$-DIP  and \cref{assu:bounding_box} such that \[{\rm Reg}(Z, {\cal I}) = \Omg(\sqrt {kT}).\]
\end{theorem}

{\new We will choose $U=\{-2\sqrt N,2\sqrt N\}\times \{-2\sqrt N,2\sqrt N\}$, which obviously \sats\ \cref{assu:bounding_box} since the minimal distance between two units is $1$.
To highlight key ideas, in this section, we assume that $k=2$. 
The extension to general $k$ is \strfwd. 

\subsubsection{Challenge} 
One may think that this lower bound follows trivially from the lower bound of the (one-by-one version of) best-expert problem (see, e.g., Section 4 of \citealt{arora2012multiplicative}).
At a high level, the $k\times T$ reward table is a random \mtx\ whose entries are i.i.d. Bernoulli with mean $1/2$. We then argue that 
\benum 
\item[a)] the expected\footnote{over both the randomness of the reward table and the policy} regret of any policy is $T/2$, and\\ 
\item[b)] by a standard anti-concentration bound, with high \prb\ (w.h.p.) there is a fixed-arm policy with total reward $T/2+ \Omg(\sqrt T)$.
\eenum

However, the reward functions constructed here are defined only for ${\bf 1}^U$ and ${\bf 0}^U$, whereas in our problem, we need to specify the reward on the entire hypercube $\{0,1\}^U$ subject to the $\psi$-DIP.
As a key step, we show that such an extension is always possible.}


{\new \begin{lemma}[Extension to the Hypercube]
\label{prop:homotopy} 
Let $G=(V,E)$ the  grid graph where $V = \{-m,\dots,m\}\times \{-m,\dots,m\}$ for some integer $m$.
Denote by $O\in V$ the origin. 
Then, for any non-increasing function $\psi:\real_+\rar\real_+$, there exists a function $f: \{0,1\}^V \rar [0,1]$ \sat ing\\
{\bf i) the boundary condition:} \[f({\bf 0}^V)=0, \quad f({\bf 1}^V)=\psi(0)-\psi(m),\quad {\rm and}\] 
{\bf ii) the $\psi$-DIP:} for any $z,z'\in \{0,1\}^V$ with $z_{B(O,r)}=z'_{B(O,r)}$ for some $r>0$, we have \[|f(z) - f(z')|\le \psi(r).\]
\end{lemma}
\proof Recall} that $d(\cdot)$ denotes the sup norm and $B(v,r)$ denotes the ball centered on $v$ with radius $r$. 
We construct $f$ in the following two steps.
\benum 
\item {\bf  Define $f$ on basis vectors.} 
For each $r=0,1,\ldots,m$, we define the {\em basis vector} ${\boldsymbol \sigma}^r \in \{0,1\}^V$ {\new whose entries are given by $\sigma^r_v = {\bf 1}(d(O,v)< r)$} for each $v\in V.$
In \parti, ${\boldsymbol \sigma}^0 = {\bf 0}^U$ and ${\boldsymbol \sigma}^r = {\bf 1}^U$. 
Define \[f\lb({\boldsymbol \sigma}^r \rb) := \psi(0)-\psi(r), \quad r=0,\dots,m.\]
\item {\bf Extend $f$ to $\{0,1\}^V$.} 
For each $z\in \{0,1\}^V$, define \[f(z) = f({\boldsymbol \sigma}^{r_\star(z)}) \quad \text{where} \quad  r_\star(z) =\max\{r\ge 0: z_{B(O,r)} = {\bf 1}_{B(O,r)}\}.\]
\eenum
{\new Note that $f({\boldsymbol \sigma}^0) = \psi(0)-\psi(0) = 0$ and $f({\boldsymbol \sigma}^m) = \psi(0)-\psi(m)$, and so (i) holds.} To show (ii), fix any $z,z'\in \{0,1\}^V$. For simplicity, we write $r_\star:= r_\star (z)$ and $r_\star'=r_\star (z')$.
W.l.o.g. we assume that $r_\star \le r_\star'.$ Consider the largest ball $B(O,\rho)$ on which $z$ and $z'$ ``agree'', that is,
\[\rho := \max\{r\ge 0: z_{B(O,r)} = z'_{B(O,r)}\}.\] 
{\new  {\bf Claim:} $\rho \ge r_\star$.} 

Assuming the claim, we conclude that
\begin{align*}
|f(z) - f(z')| &= |(\psi(0)-\psi(r_\star)) - (\psi(0)-\psi(r_\star'))| \\
&= \psi(r_\star) - \psi(r_\star') \\
&\le \psi(r_\star)\\
&\le \psi(\rho),
\end{align*} where the last \ineq\ follows since $\psi$ is non-increasing.\qed



\noindent{\bf Proof of the Claim.} Consider two cases. 
{\new If $r_\star= r_\star'$, then the claim trivially follows from the definition of $\rho$.
Now \sps\ that $r_\star'> r_\star$. Note that $r_\star, r_\star'$ are both integers, so $r_\star' \ge r_\star+1$.
By the definition of $r_\star$, there exists $v\in B(O,r_\star +1)\bs B(O,r_\star)$ such that $z_v=0$ and $z'_v=1$. Therefore, $z$ and $z'$ do not ``agree'' on $B(O,r_\star+1)$, so 
\begin{align}\label{eqn:071224}
\rho < r_\star+1.\end{align} 
On the other hand, since we assumed $r_\star< r_\star'$, we have $z_{B(O,r_\star)} =z'_{B(O,r_\star)}$, and thus \begin{align}\label{eqn:071224b}
\rho \ge r_\star.
\end{align}
Combining \cref{eqn:071224,eqn:071224b}, we have $\rho = r_\star$.}
The claim follows by combining the two cases.\qed

\subsubsection{Proof of Theorem \ref{thm:lb}}
{\new Recall that in the construction we chose $U=\{-2\sqrt N,2\sqrt N\}\times \{-2\sqrt N,2\sqrt N\}$. 
By re-scaling, w.l.o,g, let us assume that $\psi(0)=1$ and $\psi(\sqrt N) = 0$. Consider the interior units \[U_{\rm int}:= \{-\sqrt N,\sqrt N\}\times \{-\sqrt N,\sqrt N\}.\]
For each $u\in U\bs U_{\rm int}$, we define $Y_{ut}\equiv 0$ for all $t\in [T]$.

To define the reward function for the interior units, for 
each $u\in U_{\rm int}$ we apply \cref{prop:homotopy} with $m=\sqrt N$ and \[\quad V=\{u_x - m, u_x + m\}\times \{u_y - m, u_y + m\},\] where $u=(u_x,u_y)$.
Denote by $\tilde f_u: \{0,1\}^V\rar \real_+$ the function constructed in \cref{prop:homotopy}, and extend $f_u$ to $\{0,1\}^U$ so that  the function value of $u'$ is solely determined by its restriction on $V$, formally, \[f_u(u') = \widetilde f_u(u'|_V), \quad \forall u'\in\{0,1\}^U.\]
By \cref{prop:homotopy}, $\tilde f_u$ \sats\ the $\psi$-DIP, and so $f_u$ also \sats\ the $\psi$-DIP.

We are now ready to explicitly construct the reward function $Y_{ut}$:}
Consider an i.i.d. \xulie\ of Bernoulli variables $\xi_t \sim {\rm Ber}(1/2)$ where $t\in [T]$, we define 
\[Y_{ut}(z) = \frac 12 + \lb(\xi_t-\frac 12\rb)f_u(z), \quad \forall z\in \{0,1\}^U.\]
Then, by \cref{prop:homotopy}, $Y_{ut}(\cdot)$ satisfies the $\psi$-DIP.  
Moreover, for any fixed $z\in \{0,1\}^U$, we have $\ho{E}_\xi [Y_{ut}(z)] = \frac 12$.
Therefore, the expected reward of any policy is $T/2$.

Next, we show that the best fixed-arm policy has a total reward $T/2 +\Omg(\sqrt T)$ in expectation.
This requires the following standard result.

\begin{lemma}[Bernoulli Anti-concentration Bound]
\label{lem:anti_concentration}
Let $(\xi_t)_{t\in [T]}$ be i.i.d. Bernoulli variables with mean $\frac 12$, and write $\xi = \sum_{t=1}^T \xi_t$. Then, for any $s\in [0,\frac T8]$, we have
\[\ho{P}\lb[\xi\ge \frac T2 + s\rb]\ge \frac 1{15} \exp\lb(-\frac {16 s^2}T\rb)\]
\end{lemma}

To conclude, let $R_a$ be the total reward of arm $a$.
Consider the event ${\cal E} = \{R_1 \ge \frac T2 + \frac {\sqrt T}4\}$.
Since $R_1$ is the sum of $T$ i.i.d. Bernoulli's, by taking $s = \frac 14 \sqrt T$ in \cref{lem:anti_concentration}, we have $\ho{P}\lb[{\cal E}\rb]\ge \frac 1{15}.$ Therefore, 
\begin{align*}
 \ho{E} \lb[\max\lb\{R_0, R_1\rb\} -\frac T2 \rb]
&\ge \ho{E}\lb[\max\lb\{R_0, R_1\rb\} -\frac T2 \middle |\ \bar {\cal E}\rb] \cdot \ho{P}\lb[\bar {\cal E}\rb] \\
&\ge\frac 1{15} \cdot \frac{\sqrt T} 4 = \frac{\sqrt T}{60},
\end{align*}
where the first \ineq\ follows since $\max\{R_0,R_1\} \ge \frac T2$ a.s.\qed

\section{High Probability Regret Bound}
Although switchback policies can achieve optimal expected regret, a notable drawback is their disregard for $N$, preventing them from capitalizing on the market size. 
Consequently, the fluctuation of regret (as a random variable) remains unaffected by $N$, making it less appealing to risk-averse decision-makers.

To address this, we propose a policy that (i) has a $\widetilde O(\sqrt T)$ expected regret and (ii) admits a high probability (h.p.) regret that vanishes in $N$.
Specifically, for any number $T$ of rounds and confidence level $\delta>0$, the tail mass above $\widetilde \Theta(\sqrt T)$ vanishes $N\rar \infty$.
Our policy integrates the EXP3-IX policy from adversarial bandits and the idea of clustered randomization from causal inference. 
Let us begin by explaining these two basic concepts.

\subsection{Background: Implicit Exploration and Clustered Randomization}
\label{sec:prelim_mwa}
{\new Our approach integrates two ideas: implicit exploration in the EXP3-IX policy for adversarial bandits, and clustered randomization for causal inference under (spatial) interference.}

\subsubsection{The EXP3-IX Policy}
Policies for adversarial bandits often rely on {\em weights update} \citep{vovk1990aggregating,littlestone1994weighted}.
In each round, an arm is selected with a probability proportional to its weight, which is updated to incentivize choosing arms with historically high rewards.
However, with bandit feedback, we only observe the reward of the selected arm, which poses a challenge for weight updates. 

{\new The EXP3 policy addresses this by employing a reliable estimate of the rewards of {\bf all} arms. 
The most basic estimator is the following {\em importance-weighted estimator} 
\citep{auer1995gambling}:
In round $t$, let $P_{ta}$ be the \prb\ that $a$ is selected, $A_t$ be the arm selected, and $Y_t$ be the observed reward (of $A_t$), define \[\wh Y_{ta} := \frac {{\bf 1}(A_t = a)} {P_{ta}} Y_t.\]
The EXP3 policy works by combining this estimator with the multiplicative weights \alg, and achieves an optimal $\widetilde O(\sqrt {kT})$ expected regret against the best fixed arm.}

However, the regret of the EXP3 policy is prone to high variance. 
In fact, the regret can be {\em linear} in $T$ w.p. $\Omg(1)$;\footnote{This does not contradict the (expected) regret bound, since the regret against the best fixed arm can be  negative.} see Note 1 in Chapter 11 of \cite{lattimore2020bandit}.
This is essentially because the estimator in EXP3 can have a high variance.
To address this, \cite{kocak2014efficient} introduced an {\em implicit exploration} (``IX'') term $\beta>0$ in the propensity weight, which truncates the value of the estimator and consequently reduces its variance. 
{\new Formally, we define \[\wh Y_{ta} := \frac {{\bf 1}(A_t = a)} {P_{ta}+\beta} Y_t.\]
Despite the extra bias, \cite{neu2015explore} showed that judicious selection of $\beta$ leads to a good high \prb\ (h.p.) regret bound: The regret is within $\sqrt{\log 1/\delta}$ times the minimax regret, $\tilde O(\sqrt{kT})$, w.p. $1-\delta$.}



\subsubsection{Cluster-Randomization}
\cite{leung2022rate} considered inference under spatial interference, assuming that the {\em potential outcomes} (i.e., rewards, in the terminology of MAB) \sats\ the DIP.
They partition the plane uniformly into square-shaped clusters and independently assigned treatment arms to each cluster.
Under this design, their  truncated HT type estimator achieves favorable bias and variance, both vanishing in $N$.

\subsubsection{Challenges} There are two main challenges in integrating the truncated HT estimator into the EXP3-IX framework. 
First, the uniform spatial clustering in \citealt{leung2022rate} is no longer ``robust'' since some arms may have very low \prbs\ due to weight updates, leading to high variance in the estimator.
Furthermore, it is unclear how to select the IX parameter in the batched setting due to the heterogeneity across units.
{\new For example, units that may lie close to the boundary of a cluster  should have different IX \pmt s compared to those that lie in the ``interior'' of a cluster.}

{\new We explain how to tackle these two challenges in the rest of this section.
We start by defining the key component of the (truncated) HT estimator in \citealt{leung2022rate} is the exposure mapping. }

\bdefn[Exposure Mapping]
For each $u\in U$, $t\in [T]$ and $a\in [k]$, we define the {\em radius-$r$ exposure mapping} as \[X_{uta}^r(z) = \prod_{v\in B(u,r)} {\bf 1}(z_v =a),\quad \forall z= (z_u)\in [k]^U.\]
For a random vector $Z\in [k]^U$,  we define the {\em exposure \prb} as \[Q_{uta}^r(Z) = \ho{P} \lb[X_{uta}^r(Z)=1\rb]\]
\edefn

In words, the exposure mapping measures the ``reliability'' of observations from a unit. 
Specifically, it is the binary indicator for whether {\em all} units in a neighborhood are assigned the same treatment.
Therefore, to achieve favorable statistical performance, we prefer a high exposure probability. 
{\new However,  in the naive clustered randomization, the exposure \prb\  of some units can be very low, leading to a high variance in the estimate (and cumulative regret). We address this issue next.}

\subsection{Robust Randomized Partition}
\label{sec:RRC}

If $T=1$ and we aim to estimate the ATE, uniform clustering works well. 
For example, in \cite{leung2022rate}'s setting, {\new  by choosing $r<\ell/2$ where $\ell$ is the side length of the square clusters, each $r$-ball intersects at most $4$ squares.
Therefore, if we \indep ly assign each cluster an arm $a$ with \prb\ $p=p_a=\Omg(1)$ (and some other arm w.p. $1-p$),} then the exposure probability is at least $p^4 = \Omg(1)$, which is fine for estimation.

However, {\new in the presence of weight updates,} $p$ (and therefore $p^4$) can be very small.
{\new When this occurs, the exposure mapping of a unit near the boundary of a cluster is very unlikely to be $1$.}
\IOW, we almost remove all data from these units.
This results in a high bias in the estimation. 
{\new For example, \sps\ the rewards are zero everywhere except near the cluster boundaries. 
Then our estimator is $0$, but the true reward is $\Omg(1)$.}

\begin{figure}
\centering
\includegraphics[width=8cm, height=6cm]{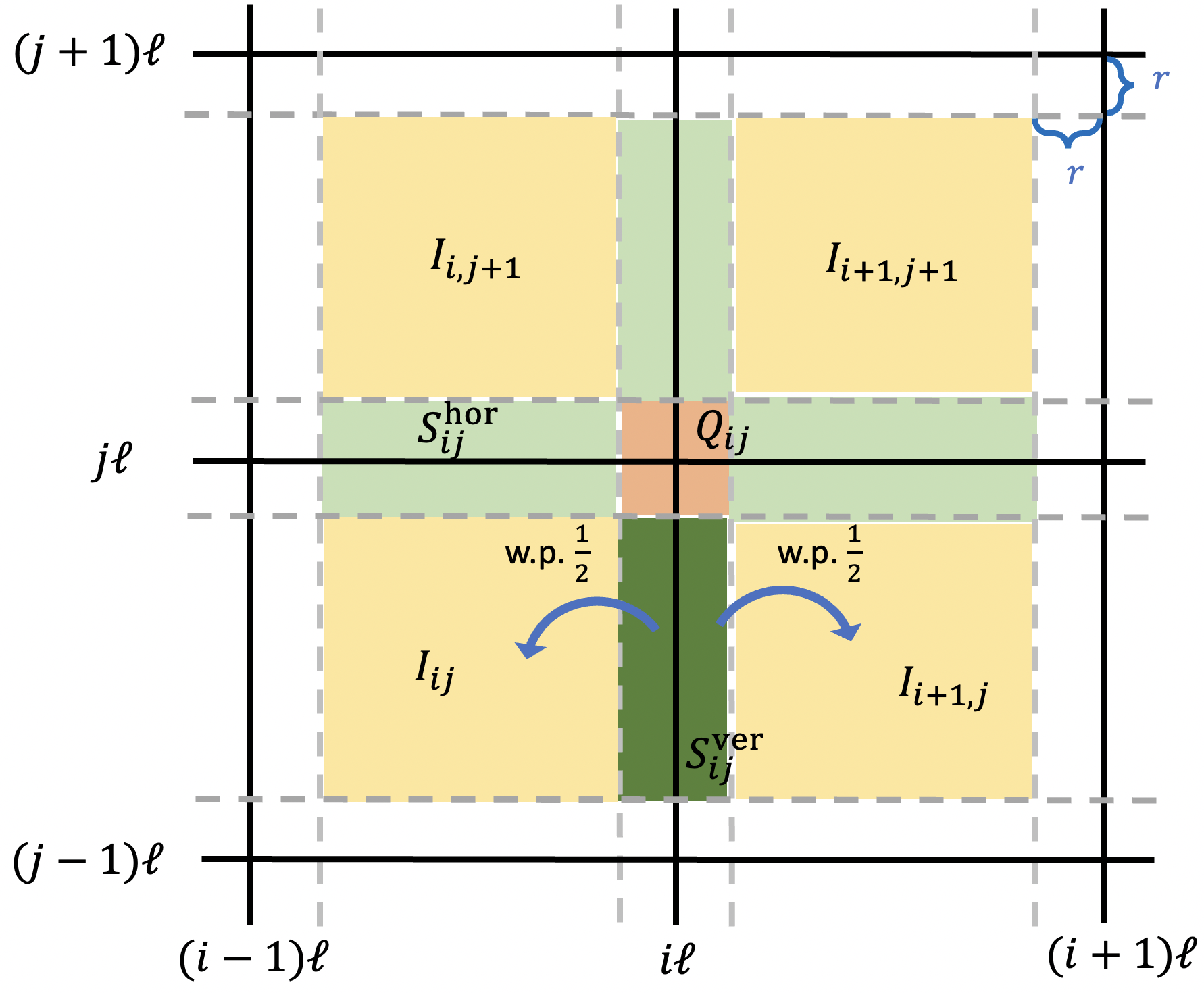}
\caption{Illustration of the RRP. The black lines are the boundary for the squares in the uniform clustering.
We color the strips and quads green and red. 
We assign each strip to one of the two neighboring clusters; see $S^{\rm ver}_{ij}$ (dark green).
Finally, assign each quad (red) to one of the four nearby clusters with equal \prbs.}
\label{fig:rrp}
\end{figure}

We address this by introducing {\em randomness} into the partition.
We will start with uniform clustering, and then randomly assign units close to the boundary to nearby clusters.
{\new To formalize this, recall from \cref{assu:bounding_box} that $U\sse [0,b_N]^2$ where $b_N =O(\sqrt N)$.}
For any $\ell,r>0$ with $r< \ell /2$, an {\em $(\ell,r)$-robust randomized partition} (RRP) $\Pi = \{C_{ij}:1\le i,j \le b_N/\ell\}$ is defined as follows; see \cref{fig:rrp} for an illustration. 

\benum
\item {\bf Assign the Interiors:} Define 
the {\em $(i,j)$-interior} as 
\[I_{ij} = \lb[(i-1)\ell + r,\ i\ell -r\rb]\times \lb[(j-1)\ell + r,\ j\ell -r\rb].\]
We assign $I_{ij}$ to $C_{ij}$ w.p.  $1$.
\item {\bf Assign the Strips:} Define the {\em vertical $(i,j)$-strip} 
\[S_{ij}^{\rm ver} = \lb[i\ell - r,\ i\ell + r\rb] \times \lb[(j-1)\ell+r,\ j\ell - r\rb],\] and the {\em horizontal $(i,j)$-strip} 
\[S_{ij}^{\rm hor} = \lb[(i-1)\ell + r,\ i\ell - r\rb] \times \lb[j\ell - r,\ j\ell + r\rb].\]
We assign each $S^{\rm ver}_{ij}$ independently to $C_{ij}$ or $C_{i+1,j}$ \unif ly. 
Similarly, assign $S^{\rm hor}_{ij}$ to independently to $C_{ij}$ and $C_{i,j+1}$ \unif ly.
\item {\bf Assign the Quads:} Define the {\em $(i,j)$-quad}  \[Q_{ij} = \lb[i\ell-r, i\ell+r\rb] \times \lb[j\ell-r, j\ell+r\rb].\] Assign it to $C_{ij}, C_{i+1, j}, C_{i,j+1}$ and $C_{i+1,j+1}$ uniformly at random. \qed
\eenum
{\new Our clustering is obtained by partitioning $U$ using an RRP. Formally, let $\{C_{ij}\}$ be an $(\ell,r)$-RRP of $[0,\sqrt N]^2$. 
We will use the clustering $\{U\cap C_{ij}\mid 1\le i,j\le b_N /\ell\}$.
By abuse of notation, let us abbreviate $U\cap C_{ij}$ as $C_{ij}$.
For each $u\in U$, denote by $C[u]\sse U$ the unique cluster that contains $u$.\footnote{breaking ties \arbly.}}
From now on 
let us fix a pair of $\ell,r\ge 0$ with $1\le \ell \le b_N$ and $2r<\ell$.
The RRP enjoys the following nice {\em robustness}.

\begin{proposition}[Robustness]\label{prop:robustness}
For any $u\in U$, we have $\ho{P}[B(u,r)\sse C[u]]= \Omg(1).$
\end{proposition}

To see this, observe that since $r<\ell/2$, the ball $B(u,r)$ intersects at most four ``regions'' (i.e., interiors, strips or quads). 
Since each strip or quad is assigned to a cluster independently, w.p. $\Omg(1)$ these regions are all assigned to the same cluster $C$. 
When this occurs, we have $B(u,r)\sse C[u]$.

The robustness boosts the exposure probability.
In fact, our policy (to be defined soon) maintains a weight for each arm and assigns a random arm to each cluster \indep ly according to the weights.
{\new Crucially, when $P_{ta}$ is small, the exposure \prb\ under our RRP is $\Omg(P_{ta})$, which is much greater than the exposure \prb\ $(P_{ta})^4$ under the uniform design \citep{leung2022rate}.}
We will soon see how this helps reduce the variance of our estimator, which we define next.

{\new \begin{remark} Our approach may resemble that of \cite{ugander2023randomized}.
Their {\bf randomized graph clustered randomization} (RGCR) is a two-stage procedure: First, find a good clustering (e.g., $3$-net).
Then, sort these clusters by cardinality and pair up the $(2j+1)$-st and $(2j+2)$-nd clusters, $j=0,1,\dots$.
Finally, for each pair, randomly assign the treatment to exactly one of the two clusters. 
By doing this, the number of treatment and control clusters differs by at most $1$, reducing the variance of the estimator. 

Our approach adopts a fundamentally different idea. Unlike the RGCR, our clustering itself is random. 
In \parti, our approach randomizes the ``interiorness'' of a unit, so that every fixed unit has a reasonable \prb\ to lie within the interior of its cluster, thereby having an $\Omg(1)$ exposure \prb. 
In contrast, the RGCR does not have this property. 
\end{remark}}

\subsection{HT-IX Estimator and Our Policy}\label{sec:THT}
In adversarial bandits, we only observe the rewards of the selected arm. Therefore, the key to designing a good policy is finding a good estimator for the arms' reward. 
More explicitly, in the MABI problem, we need to find a good estimator for the mean rewards $\{\bar Y_t (a\cdot {\bf 1}^U)\}_{a\in [k]}$, where we recall that $\bar Y_t(Z) = \frac 1N \sum_{u\in U} Y_{ut}(Z)$ for any $Z\in [k]^U$.

Ideally, a good estimator should have both low bias and low variance.
\citet{neu2015explore} showed that incorporating an additional IX parameter into the propensity weight results in a favorable high-probability regret bound.
In the MABI problem, the propensity weights $Q_{uta}^r$ vary across units.
However, a uniform IX parameter suffices for our result.

\bdefn[Horvitz-Thompson-IX Estimator] 
Fix an implicit exploration (IX) \pmt\ $\beta \in [0,\frac 12)$. {\new  
\Sps\ $Z_t\in [k]^U$ (the ``design'') is drawn from a \distr\ $\cal D$.\footnote{More concretely, we will later choose $\cal D$ to be a clustered-level randomization, that is, assign a random arm to each cluster drawn with a \prb\ proportional to its weight.} 
Denote \[Q_{uta}^r := \ho{P}_{Z_t \sim \cal D}(X_{uta}^r(Z_t) =1).\]
For} each $t\in [T]$, $a\in [k]$, the {\em Horvitz-Thompson-IX} (HT-IX) estimator is 
\[\wh Y_t(a) := \frac 1N \sum_{u\in U} \frac {{\bf 1}(X_{uta}^r(Z_t) =1)}{Q_{uta}^r+\beta} Y_{ut}(Z_t).\]
\edefn

\brmk Our HT-IX estimator ``interpolates'' between the estimators in \citealt{leung2022rate} and \citealt{neu2015explore}:
When $\beta = 0$, it is the HT estimator in \citealt{leung2022rate}, and
when $N=1$, it becomes the estimator in the EXP3-IX policy of \citet{neu2015explore}.
\ermk 
We are now ready to describe our policy.
It involves two parameters: 
The learning rate $\eta\in (0,1)$, which controls how quickly we discount past data, and the IX parameter $\beta\in [0,\frac 12)$ truncates the HT-IX estimator by $1/\beta$.
In each round, we {\bf independently} generate an $(\ell,r)$-RRP. 
Then, we randomly assign an arm to each cluster independently, using the \distr\ determined by the weights.
Finally, for each arm, we use the HT-IX estimator to estimate the counterfactual reward that we could have earned if we assigned it to {\bf all} units in this round.
We update the arm weights using the estimated rewards. 
A formal definition is given in \cref{alg:batched_exp3}.

\renewcommand{\algorithmiccomment}[1]{\hfill \textit{\textcolor{blue}{// #1}}}

\begin{algorithm}[t!]
\caption{EXP3-HT-IX Policy}
\label{alg:batched_exp3}
\begin{normalsize} 
\begin{algorithmic}[1]
\STATE{Input: \\
$\eta \in (0,1)$: learning rate, \\
$\beta\in [0,\frac 12)$: IX parameter for the HT-IX estimator, \\
$(\ell,r)$: parameters for the RRP.}
\STATE{$W_{ta}\lar 1$ for each $a\in [k]$}
\FOR{$t=1,\dots,T$}
\STATE $W_t \lar \sum_{a\in [k]} W_{ta}$ \COMMENT{Total weights}  
\STATE{For each arm $a\in [k]$, let $P_{ta} \lar W_{ta} / W_t$}
\STATE{Randomly generate $\Pi_t$, an $(\ell,r)$-RRP}
\FOR{cluster $C\in \Pi_t$}
\STATE{Draw an arm $Z_{Ct}$ using $(P_{ta})_{a\in [k]}$}
\FOR{$u\in C$} 
\STATE{$Z_{ut}\lar Z_{Ct}$} \COMMENT{Assign arm $A_t$ to all units in $C$}
\ENDFOR
\ENDFOR 
\STATE{Observe the rewards $\{Y_{ut}(Z_t)\}_{u\in U}$}
\FOR{$a\in [k]$}
\STATE $\wh Y_t(a) \lar \frac 1N \sum_{u\in U} \frac {{\bf 1}(X_{uta}^r(Z_t) =1)}{Q_{uta}^r+\beta} Y_{ut}(Z_t)$ \COMMENT{Apply the HT-IX estimator}
\STATE $W_{t+1, a} \lar e^{\eta \wh Y_t(a)} W_{ta}$ \COMMENT{Weight update}
\ENDFOR
\ENDFOR 
\end{algorithmic}
\end{normalsize} 
\end{algorithm}
 
\subsection{A High-Probability Regret Bound}\label{sec:statement_of_hp_regret}
We introduce some notation before stating our results.
Given a policy $Z=(Z_t)_{t\in [T]}$, we denote by \[R= \sum_{t=1}^T \frac 1N \sum_{u\in U} Y_{ut}(Z_t)\] the total reward per unit.
To compare against a fixed arm $a$, we also define \[R_a := \sum_{t=1}^T \bar Y_t (a)\quad \text{where}\quad \bar Y_t (a):=\frac 1N \sum_{u\in U} Y_{ut}(a\cdot {\bf 1}^U).\]

\begin{theorem}[Upper Bound]
\label{thm:main}
Fix any IX \pmt\ $\beta\in (0,\frac 12)$ and learning rate $\eta\in (0,1)$ in \cref{alg:batched_exp3}. 
Then, for any arm $a^*\in [k]$ and confidence level $\delta \in (0,1)$, we have 
\[R-R_{a^*} = A+B+C \quad {\rm w.p.}\ 1-\delta\] where
\[\quad A\ls \frac {\log k}\eta + \eta kT, \quad B\ls \lb(\frac 1\beta  + \eta kT\rb) {\new \frac{\ell^2}N }\log \frac 1\delta, \quad \text{and}\quad C\ls \beta kT + \frac {rT}\ell + \eta T\frac k\beta \psi(r)^2 +\frac T {\beta \ell^2} \psi(r).\]
\end{theorem}

{\new 
We will soon see (from the corollaries in the next subsection) that with a suitable choice of the \pmt s, $A$ becomes the minimax expected regret $\tilde O(\sqrt{kT})$, $B$ bounds the tail mass of the regret, which vanishes as $N\rar \infty$, and $C$ corresponds to the lower order terms.}

\subsection{Implications of Theorem \ref{thm:main}}
{\new We illustrate \cref{thm:main} via several corollaries. 
To minimize $A$, choose $\eta=\sqrt{\log k /kT}$. 
The choice for $\beta$ is more involved. 
With some foresight, let us choose $\beta =\sqrt {\frac {\ell^2}{kNT} \log \frac 1\delta}$. 
To highlight the excess beyond the ``necessary'' regret, we denote ${\rm Reg}_{\rm OPT} = \sqrt {kT\log k}$ as the order-optimal expected regret.}

We first consider the no-interference setting. In this case, our problem is equivalent to the multiple-play (i.e., play $N$ arms in each round) variant of adversarial bandits. 
{\new Since there is no interference, we will choose the singleton clustering (i.e., where each unit alone is a cluster). 

This clustering can be realized as an $(\ell,r)$-RRP for suitably small $\ell$ and $r$.
In fact, recall from \cref{assu:bounding_box} that $d(u,v)\ge 1$ for any $u,v\in U$. 
Therefore, if we partition the bounding box $[0,b]^2$ uniformly into squares of sufficiently small side lengths $\ell$, then each square contains at most one unit in $U$. 
Thus, the $(\ell,r)$-RRP is just the singleton clustering (whenever $r< \ell /2$). 
Finally, noting that $\psi(r) = 0$ for any $r>0$
and hence $C\ls \beta kT+o(1)$ as $r\rar 0^+$, we obtain the following.

\bcoro[No Interference]
\label{coro:no_inter}
\Sps\ $\psi(x) = {\bf 1}(x=0)$. 
Then, there exists $\ell,r$ such that the regret of the EXP3-IX-HT with the $(\ell,r)$-RRP \sats\ 
\begin{align}\label{eqn:052224}
R-R_{a^*} \ls \lb(1+\sqrt{\frac 1N \log \frac 1\delta}\rb){\rm Reg}_{\rm OPT} \quad {\rm w.p.}\ 1-\delta
\end{align}
for every $a^*\in [k]$ and $\delta\in (0,\frac 12)$.
\ecoro

To better understand, recall that for (the one-by-one version of) adversarial bandits, the regret of EXP3-IX is $(1+\sqrt { \log 1/\delta})\cdot {\rm Reg}_{\rm OPT}$; see Chapter 12 in \citealt{lattimore2020bandit}.
To see the difference, take $N=T$ and $\delta=N^{-\Omega(1)}$, this bound is $O(\log N \cdot {\rm Reg}_{\rm OPT})$, while \eqref{eqn:052224} is $O({\rm Reg}_{\rm OPT})$ for any $T$.}

{\new Another basic setting is {\em $\kappa$-neighborhood} interference (see, e.g., \citealt{leung2022rate,bojinov2023design}). 
Here, the reward (``potential outcome'') of a unit depends only on the treatment assignment of the units within distance $\kappa>0$. 
Formally, the interference function is $\psi(x) = {\bf 1}(\kappa>x)$. In this case, by selecting $r=\kappa$, the $\psi(r)$ terms in $C$ become $0$.}

\bcoro[$\kappa$-Neighborhood Interference]\label{coro:h-nbhd}
\Sps\ $\psi(x) := {\bf 1}(\kappa>x)$ for some $\kappa>0$. 
Then, with $r=\kappa$ and $\ell =\kappa \sqrt T$, for any $a^*\in [k]$ and $\delta\in (0,\frac 12)$, we have
\[ R-R_{a^*} \ls\lb(1+ \kappa\sqrt{\frac TN \log \frac 1\delta}\rb){\rm Reg}_{\rm OPT} \quad {\rm w.p. }\ 1-\delta.\]
\ecoro

Finally, we consider the setting where $\psi(r)$ follows the {\em power law}.
This setting encompasses many fundamental settings, including the celebrated Cliff-Ord spatial autoregressive model
\citep{cliff1973spatial}, which posits that each unit's outcome is linear in its neighbors' treatments; see detailed explanation in \citealt{leung2022rate}.

\bcoro[Power-law Interference]
\label{coro:power_law}
\Sps\ $\psi(r) =O(r^{-c})$ for a constant $c\ge 1$. 
{\new Consider an $(\ell,r)$-RRP with $m=\min \{(N/T)^{\frac {2+c}{3+c}},\ N^{\frac {2c}{2c+1}} T^{-\frac {2c-1}{2c+1}}\}$ clusters (and hence $\ell= \sqrt {N/m}$) and $r = \ell /\sqrt T$.}
Then, for any $a^*\in [k]$ and $\delta\in \frac 12$, we have
\begin{align}\label{eqn:051824}
R-R_{a^*} \ls \lb(1+\sqrt{ \frac{\ell^2}N \log \frac 1\delta}\rb) {\rm Reg}_{\rm OPT} \quad {\rm w.p.}\ 1-\delta.
\end{align}
\ecoro

For example, when $c=2$, we have $m =(N/T)^{4/5},$ and thus
$\eqref{eqn:051824} = \widetilde O (\sqrt {kT}) + k \frac{T^{9/10}}{N^{2/5}} \sqrt {\log \frac 1\delta}.$

\subsection{Discussion: Interpretation in VaR}
\begin{figure}[h]
\centering
\begin{minipage}{.45\linewidth}
  \centering
  \includegraphics[width=0.9\linewidth]{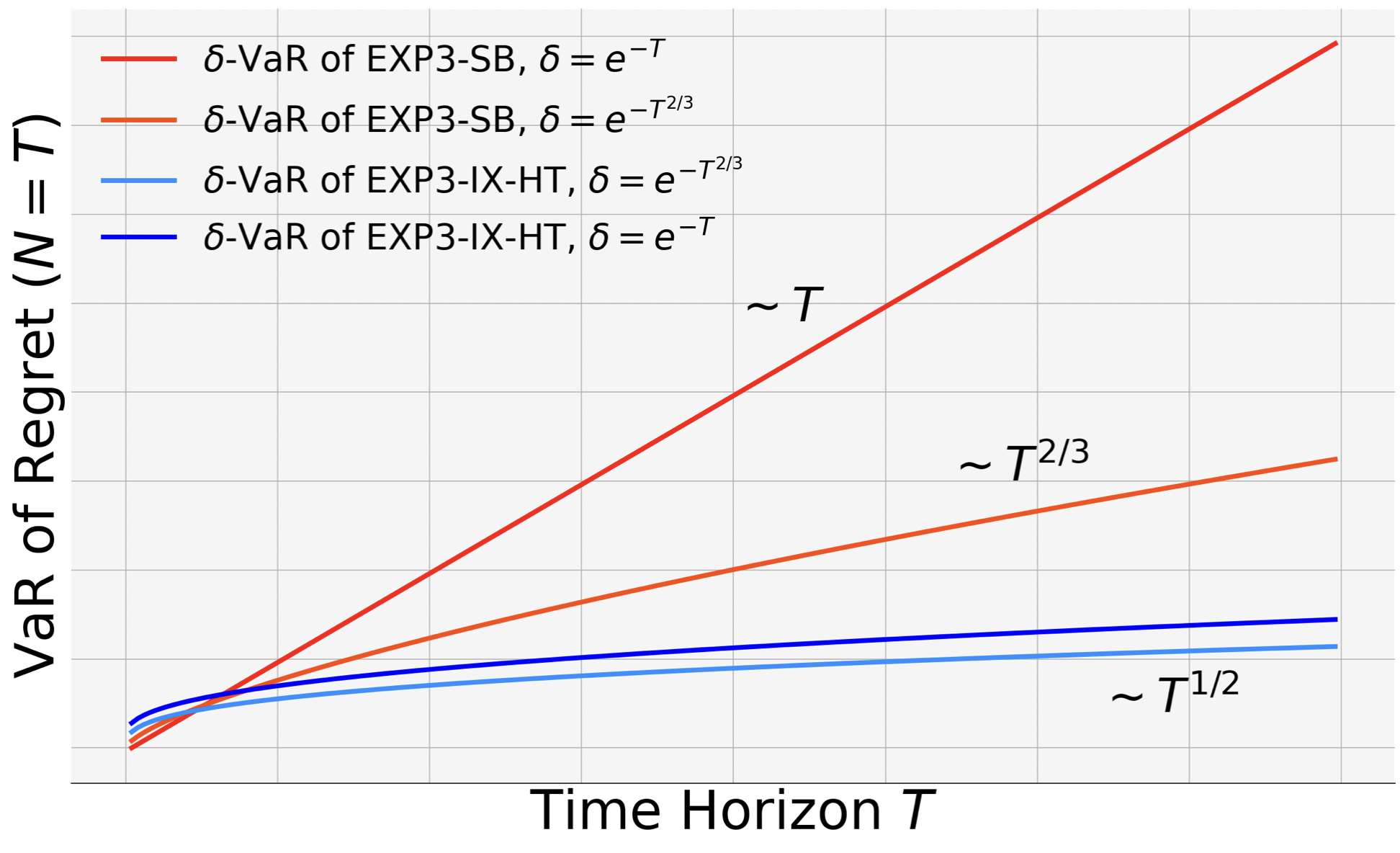}
  \captionof{figure}{VaR of Regret. We visualize the $\delta$-VaR of regret for $\delta = e^{-T}$and $\delta = e^{-T^{2/3}}$ \resp. 
  Here we set $c=1/2$. 
  Our cluster-randomization based policy has a much lower VaR.
  }
  \label{fig:VaR}
\end{minipage}
\hspace{0.5cm}
\begin{minipage}{.45\linewidth}
  \centering
  \includegraphics[width=0.9\linewidth]{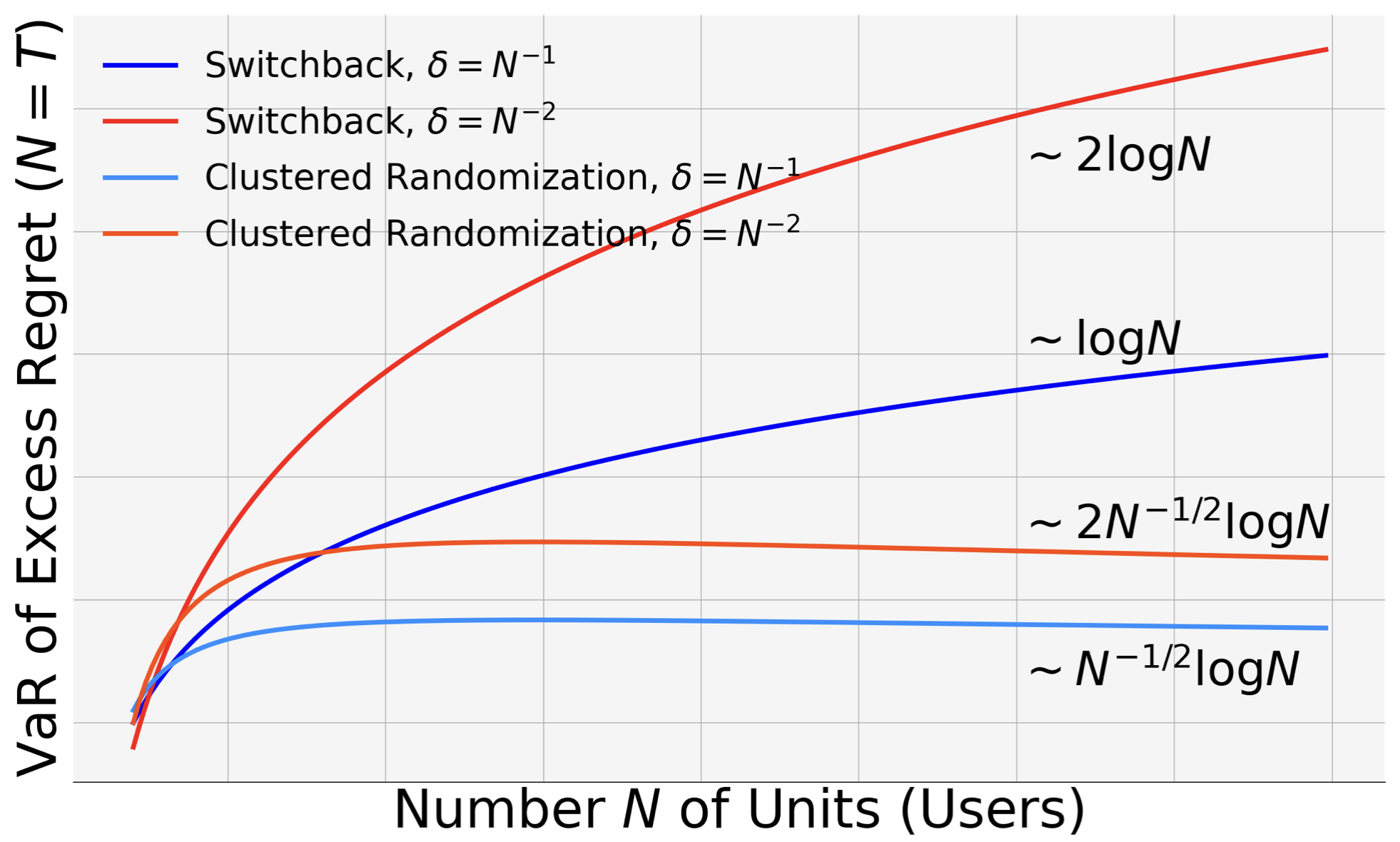}
  \captionof{figure}{VaR of Excess Regret: The figure visualizes the excess regret ${\rm Reg} -{\rm Reg}_{\rm OPT}$ that we can guarantee w.p. $1-\delta$.}
  \label{fig:excess regret}
\end{minipage}
\end{figure}

{\new By taking $\eta\sim \sqrt {1/kT}$ in 
\cref{thm:main}, our EXP3-IX-HT policy also achieves the optimal $\tilde O(\sqrt{kT})$ expected regret.} 
However, our policy has a substantially lower tail risk compared to any switchback policy.
To better illustrate, for concreteness, let us take $N=T$.
Denote by ${\rm Reg}_{\rm SB}$ and ${\rm Reg}_{\rm CR}$ the regret (as \rv s) of any SwitchBack policy and our EXP3-IX-HT policy (where ``CR'' means ``Clustered Randomization''). 
Then, by \cref{thm:main}, 
\begin{align}\label{eqn:051924}
\ho{P}\left({\rm Reg}_{\rm SB}>{\rm Reg}_{\rm OPT} + \log\frac 1\delta\right) \le \delta\quad \text{and}\quad \ho{P} \left({\rm Reg}_{\rm CR}> {\rm Reg}_{\rm OPT} +N^{-c}\log\frac 1\delta\right) \le\delta 
\end{align}
for any $\delta>0$, where $c>0$ is a constant depending on $\psi$. 
To highlight the tail mass, \cref{eqn:051924} can be rewritten as \begin{align}\label{eqn:052324}
\ho{P}\left({\rm Reg}_{\rm SB}-{\rm Reg}_{\rm OPT} >\tau\right) \le e^{-\tau}\quad\text{and}\quad \ho{P}\left({\rm Reg}_{\rm CR}-{\rm Reg}_{\rm OPT} >\tau\right)\le e^{- N^c\tau}
\end{align}
for any $\tau>0$.

To see why ${\rm Reg}_{\rm CR}$ is more ``robust'', take $\tau=T^{2/3}$ and let $T$ range from $10$ to $50$. 
Then, the first probability in \cref{eqn:052324} ranges from 2.7\% to 0.01\%, while the second is {\bf astronomically} small. 
For example, if $N=T$ and $c=1/2$, it ranges from $10^{-14}$ to $10^{-39}$.

More generally, consider $\delta = e^{-\alpha T}$ where $\alpha>0$. Then, the $\delta$-VaR are respectively $$V_{\rm SB}= {\rm Reg}_{\rm OPT} + T^\alpha\quad \text{and}\quad V_{\rm CR}={\rm Reg}_{\rm OPT}  +\frac {T^\alpha}{\sqrt N}\sim \sqrt T,$$ where ``$\approx$'' holds when $N\gg T$, which is true in many real-world sceanrios. 
When $\alpha>1/2$, the first bound is asymptotically larger.

\section{Proof of \cref{thm:main}}\label{sec:UB_pf}
Recall that the HT-IX estimator is \[\wh Y_t(a) := \frac 1N \sum_{u\in U} \frac {{\bf 1}(X_{uta}^r(Z_t) =1)}{Q_{uta}^r+\beta} Y_{ut}(Z_t).\]
We first decompose the regret using the following fake reward. 
\bdefn[Fake Reward]
For each arm $a\in [k]$, we define \[\wh R_a =\sum_{t\in [T]} \wh Y_t(a) \quad \text{and} \quad \wh R = \sum_{t,a}   P_{ta} \wh Y_t(a).\]
\edefn 

{\new In words, $\wh R_a$ is \apx ly the total reward of always choosing arm $a$, where the true reward $\bar Y_t(a) = \frac 1N\sum_{u\in U} Y_{ut}(a\cdot {\bf 1}^U)$ is replaced with the HT-IX estimator $\wh Y_t(a)$. Similarly, $\wh R$ is a \apx ly the total reward of our policy. This is because the expected reward of our policy in round $t$ is \apx ly $\sum_a P_{ta} \bar Y_t(a)$, and $\wh Y_t(a)$ is close to $\bar Y_t(a)$ since it is a good estimator.

Let us decompose the regret using the fake rewards. Recall that $R$ is the total reward of our EXP3-IX-HT policy, and that for any $a^*\in [k]$, $R_{a^*}$ is the reward of the fixed-arm policy at $a^*$.} Then,
\begin{align}\label{eqn:decomp}
    R - R_{a^*} = (R-\wh R) + (\wh R - \wh R_{a^*}) + (\wh R_{a^*} - R_{a^*}).
\end{align}
Next, we bound each of these three terms in a subsection separately.

\subsection{Bounding the First Term $R-\wh R$}

\begin{lemma}[Bounding $R-\wh R$]\label{lem:1st_term}
It holds that $R - \wh R \le \frac {4rT}\ell + \beta \sum_{a\in [k]} \wh R_a.$
\end{lemma}
\proof {\new We begin by further decomposing $R$ and $\wh R$, allowing us to compare each term individually in the subsequent analysis. 
Let us write \[R=\sum_{t=1}^T R_t  \quad \text{where}\quad R_t = \frac 1N\sum_{u\in U}Y_{ut}(Z_t),  \quad \text{and}\quad \wh R =\sum_{t=1}^T \wh R_t \quad \text{where}\quad \wh R_t = \sum_{a\in [k]} P_{ta} \wh Y_t(a).\]
Now,} fix any $t\in [T]$.  Then, {\new 
\begin{align} 
\label{eqn:071024}
& \quad 
N(R_t -\wh R_t) \notag\\
&= \sum_u Y_{ut}(Z_t) - \sum_{u} \sum_a \frac{P_{ta} {\bf 1}(X_{uta}^r = 1)}{Q_{uta}^r +\beta} Y_{ut}(Z_t)\notag\\
&= \sum_u \lb({\bf 1}(X_{uta}^r=0\ \forall a\in [k]) +\sum_a {\bf 1}(X_{uta}^r = 1)\rb) Y_{ut}(Z_t) - \sum_{u} \sum_a \frac{P_{ta} {\bf 1}(X_{uta}^r = 1)}{Q_{uta}^r +\beta} Y_{ut}(Z_t)\notag\\
&\le \sum_u {\bf 1}\lb(X_{uta}^r=0\ \forall a\in [k]\rb)
+ \sum_{u,a} {\bf 1}(X^r_{uta} =1) Y_{ut}(Z_t)- \sum_{u,a} \frac{P_{ta}{\bf 1}(X^r_{uta}=1) Y_{ut}(Z_t)}{Q_{uta}^r+\beta},
\end{align}
where the \ineq\ is because $Y_{ut}(\cdot)\le 1$.
To proceeds, we make two observations.
First, for any $u\in U$, the exposure mappings $X_{uta}^r$ can be all $0$ {\bf only} when $u$ lies close to $\partial C[u]$, the boundary of the cluster that contains $u$. Formally, 
\[{\bf 1}\lb(X_{uta}^r=0\ \forall a\in [k]\rb) \le {\bf 1}(d(u,\partial C[u]) \le r).\]
Second, we note that \[Q_{uta}^r \le \ho{P}\lb[Z_{C[u],t} = a\rb]= P_{ta}.\] Combining, we obtain} 
\begin{align}\label{eqn:071024b}
\eqref{eqn:071024} \le {\bf 1}\lb(d(u,\partial C[u]) \le r\rb) + \sum_{u,a} {\bf 1}(X^r_{uta} =1) Y_{ut}(Z_t)- \sum_{u,a} \frac{P_{ta}\cdot {\bf 1}(X^r_{uta}=1)}{P_{ta} + \beta}Y_{ut}(Z_t).
\end{align}
Note that in each cluster, there are at most $4r\ell$ units within a distance of $r$ to $\partial C[u]$.
Since there are $N/\ell^2$ clusters, we have \[\sum_{u\in U} {\bf 1}(d(v,\partial C[u]) \le r) \le 4r\ell \cdot \frac N{\ell^2} = \frac {4rN}\ell.\]
It follows that
\begin{align*}
\eqref{eqn:071024b} &\le \frac {4rN}\ell + \sum_{u,a} \frac \beta {P_{ta} + \beta} {\bf 1}(X^r_{uta}=1) Y_{ut}(Z_t)\\
&\le \frac {4rN}\ell + \beta N \sum_{a\in [k]} \lb(\frac 1N \sum_{u\in U} \frac{{\bf 1}(X_{uta}^r = 1)} {Q_{uta}^r+\beta} Y_{ut}(Z_t)\rb)\\
& = \frac {4rN}\ell + \beta N \sum_{a\in [k]} \wh Y_t(a),
\end{align*}
where \ineqs\ follows again from $Q_{uta}^r \le P_{ta}$. 
Therefore, 
\[N\lb (R - \wh R\rb) = N\sum_{t=1}^T \lb(R_t -\wh R_t\rb) \le \frac {4rNT}\ell + N\beta \sum_{a\in [k]} \wh R_a,\]
and the lemma follows by dividing both sides by $N$.  \qed 

\subsection{Bounding the Second Term $\wh R - \wh R_{a^*}$}
We first derive an upper bound on $\wh R -\wh R_{a^*}$ by following the analysis of the EXP3 policy.
The proof of the following can be found in the analysis of the EXP3 policy; see Equation (11.13) of \cite{lattimore2020bandit}.

\begin{lemma}[EXP3-style analysis]\label{lem:EXP3_style}
Fix any $\eta\in (0,1)$. Then, for any $a^*\in [k]$,
\[\wh R -\wh R_{a^*} \le \frac {\log k}\eta  + \eta \sum_{t,a} P_{ta} \wh Y_t(a)^2\quad \text{a.s.}\] 
\end{lemma}

We next show that $\wh Y_t(a)$ is highly concentrated around $\bar Y_t(a)$, with a tail mass that vanishes in $N$. 
This is done by a careful analysis based on the Chernoff and Bernstein \ineqs.
We first introduce some basic tools. 

\begin{theorem}[Bernstein Inequality for i.i.d. Sum]\label{thm:bernstein}
Let $\{X_i\}_{i=1,\dots,n}$ be independent mean-zero \rv s with $|X_i|\le M$  a.s. where $M>0$ is a constant. 
Then, for any $t>0$, \[\ho{P}\lb[\sum_{i=1}^n X_i \ge t \rb] \le \exp\lb(-\frac {t^2}{\sum_{i=1}^n \ho{E} [X_i^2] + \frac 13 Mt}\rb).\]
\end{theorem}

\begin{theorem}[Chernoff Inequality  for i.i.d. Sum of Bernoulli's]\label{thm:chernoff}
\Sps\ $X_1,\dots,X_n\sim {\rm Ber}(p)$ are i.i.d. \rv s and $\bar \xi = \frac 1n \sum_{i=1}^n X_i$. Then, for any $\eps>0$, we have \[\ho{P}\lb[\overline \xi \ge (1+\eps)\rb]\le \exp\lb(-\frac 13 \eps^2 np\rb).\]
\end{theorem}

\begin{lemma}[Deviation of Bernoulli Sum]\label{lem:iid_mean}
\Sps\ $\delta, p\in (0,1)$ and $X_1,\dots,X_n$ are i.i.d. ${\rm Ber}(p)$ \rv s. \\
(1) \Sps\ $p\ge \frac 1n$. Then with probability $1-\delta$, we have \[\frac 1n \sum_{i=1}^n X_i\le p+ \sqrt {\frac {p \log\frac 1\delta}n}.\]
(2) \Sps\ $p< \frac 1n$, then w.p. $1-\delta$, \[\frac 1n \sum_{i=1}^n X_i \ls \sqrt{\log \frac 1\delta} \cdot \frac 1n.\] 
\end{lemma}
\proof {\bf Part 1:} \Sps\ $p\ge \frac 1n$.
We will apply Bernstein's \ineq\ on $(X_i -p)$. 
Since $X_i\in [0,1]$ a.s. and $\ho{E} [(X_i -p)^2] =p(1-p) \le p$, by taking $M=1$ in \cref{thm:bernstein}, we have
\begin{align}\label{eqn:112623b}
\ho{P}\lb[\sum_{i=1}^n X_i \ge np + t \rb] \le \exp\lb(-\frac {t^2}{\sum_{i=1}^n \ho{E} [(X_i-p)^2] + \frac 13 t}\rb) \le \exp\lb(-\frac {t^2}{np + \frac t3}\rb).
\end{align}
for any $t>0$. Let us choose $t=\sqrt {2np \log \frac 1\delta}$.
Since $p\ge \frac 1n$, we have $np > t$.
\IFT\ \[ \eqref{eqn:112623b}\le \exp\lb(-\frac{t^2 }{2np}\rb) \le \delta.\]

\noindent{\bf Part 2:} \Sps\ $p< \frac 1n$. It suffices to consider i.i.d. $\wt X_i\sim {\rm Ber}(1/n)$ since $X_i$ stochastically dominates $X_i$. 
By the Chernoff bound (\cref{thm:chernoff}), for any $\eps>0$,   \[\ho{P}\lb[\frac 1n \sum_{i=1}^n \tilde X_i \ge (1+\eps) \cdot \frac 1n\rb]\le \exp\lb(-\frac 13 \eps^2 n \cdot \frac 1n\rb).\]
In particular, for $\eps= \sqrt{3\log \frac 1\delta}$, the above bound becomes $\delta$. 
\qed

Next, we combine the above and bound $\sum_{t,a} P_{ta} \wh Y_t(a)^2$.




\begin{lemma}[Bounding the Squared Terms]\label{lem:3rd_term} For any $\delta \in (0,1)$, we have \[\sum_{t,a} P_{ta}\wh Y_t(a)^2 \le 512 \lb(1+ \frac {k\ell^2}N \log \frac 1\delta + \frac {k}\beta \psi(r)^2\rb) T \quad {\rm w.p.\ } 1-\delta.\]
\end{lemma}
\proof Denote by $m = N/\ell^2$ the number of clusters. 
By the definition of $\wh Y_{t}(a)$, for any $t\in [T],a\in [k]$,
\begin{align}\label{eqn:012824}
\wh Y_t(a) = \frac 1N \sum_{u\in U} \frac{{\bf 1}(X^r_{uta}=1)}{Q_{uta}^r + \beta} Y_{ut}(Z_t) = \frac 1m \sum_{C\in \Pi} \lb(\frac 1{\ell^2} \sum_{u\in C} \frac{{\bf 1}(X^r_{uta}=1)Y_{ut}(Z_t)}{Q_{uta}^r +\beta} \rb).
\end{align}
By the $\psi$-DIP, if $X_{uta}^r=1$, all units in $B(u,r)$ are assigned $a$, so
$|Y_{ut}(Z_t) - Y_{ut}(a\cdot {\bf 1}^U)|\le \psi(r).$
Thus, \[{\bf 1}(X^r_{uta} =1) \cdot Y_{ut}(Z_t) \le {\bf 1}(X^r_{uta}(Z_t)=1) \cdot \lb(Y_{ut}(a\cdot {\bf 1}^U) + \psi(r)\rb).\] 
\IFT\
\begin{align}
\label{eqn:012824b}
\eqref{eqn:012824} \le \frac 1m \sum_{C\in \Pi} \lb(\frac 1{\ell^2} \sum_{u\in C}\frac{{\bf 1}(X^r_{uta}=1)(Y_{ut}(a\cdot {\bf 1}^U) +\psi(r))}{Q_{uta}^r +\beta} \rb).
\end{align}
Moreover, by \cref{prop:robustness}, we have $Q_{uta}^r \ge\frac 18 P_{ta}$, so
\begin{align}\label{eqn:112223}
\eqref{eqn:012824b} & \le \frac 1m \sum_{C\in \Pi} \lb(\frac 1{\ell^2} \sum_{u\in C}\frac{{\bf 1}(X^r_{uta}=1)(Y_{ut}(a\cdot {\bf 1}^U) +\psi(r))}{\frac 18 (P_{ta} +\beta)} \rb)\notag\\
&\le \frac 8{P_{ta}+\beta} \lb(\frac 1m \sum_{C\in \Pi} \frac 1{\ell^2} \sum_{u\in C}{\bf 1}(X_{uta}^r=1)\lb(Y_{ut}(a\cdot {\bf 1}^U) + \psi(r)\rb)\rb)
\end{align}
Note that the cardinality of each cluster \sats\ $|C|\le 2\ell^2$, assuming that $r\le \ell/2$. 
So, writing \[\bar Y_{Ct} (a\cdot {\bf 1}^U):=\frac 1{|C|}\sum_{u\in C} Y_{ut}(a\cdot {\bf 1}^U),\] we obtain
\begin{align}\label{eqn:0714124}
\eqref{eqn:112223}&\le \frac 8{P_{ta}+\beta} \lb(\frac 1m \sum_{C\in \Pi} \frac 2{|C|} \sum_{u\in C}{\bf 1}(X_{uta}^r=1)\lb(Y_{ut}(a\cdot {\bf 1}^U) +\psi(r)\rb)\rb)\notag\\
&\le \frac {16}{P_{ta}+\beta} \lb(\psi(r) + \sum_{\kappa \in \{0,1,2\}^2} \frac 1m \sum_{C:\chi(C)=\kappa} {\bf 1}(Z_{Ct}=a)\cdot \bar Y_{Ct} (a\cdot {\bf 1}^U)\rb).
\end{align}
Note that for each color $\kappa$, the above sum involves $\frac 19 m$ independent \rv s. 
So by \cref{lem:iid_mean},  w.p. $1-\delta$ it holds that
\[\frac 1{\frac 19 m} \sum_{C:\chi(C)=\kappa} {\bf 1}(Z_{Ct}=a)\cdot \bar Y_{Ct}(a\cdot {\bf 1})\le P_{ta} \bar Y_t(a) + \sqrt{\frac {P_{ta} \bar Y_t(a) + \log \frac 1\delta}{\frac 19 m}}\]
Combined with \cref{eqn:0714124}, we conclude that 
\begin{align*}
\sum_{t,a} P_{ta} \widehat Y_t(a)^2 & \le \sum_{t,a} P_{ta} \lb(\frac{16}{P_{ta}+\beta} \lb(P_{ta} \bar Y_t(a) + \sqrt{\frac{P_{ta} \bar Y_t(a) + \log \frac 1\delta}{\frac 19 m}} + \psi(r)\rb)\rb)^2\\
&\le 256 \sum_{t,a} \frac{P_{ta}}{(P_{ta} + \beta)^2} \cdot 2\lb(P_{ta}^2 \bar Y_t(a)^2 + \frac{P_{ta} \bar Y_t(a) + \log \frac 1\delta}{\frac 19 m} +\psi(r)^2\rb)\\
&\le 512 \lb(\sum_{t,a}P_{ta} + 9\sum_{t,a} \frac{\log \frac 1\delta}m + \sum_{t,a} \frac{\psi(r)^2}{P_{ta}+\beta}\rb)\\
&\le 512\lb( T + 9\frac{k\ell^2 T}N \log \frac 1\delta + \frac {kT}\beta \psi(r)^2\rb),
\end{align*}
where the second \ineq\ follows since for any $a,b,c\in \real$, we have $(a+b+c)^2\le 2(a^2 + b^2 + c^2)$.\qed

\subsection{Bounding the Third Term $\wh R_a - R_a$}
We need the following tool, which is stated as Lemma 12.2 in \citealt{lattimore2020bandit}.

\begin{proposition}[One-sided Cramer-Chernoff Bound]
\label{prop:cramer_chernoff} 
Let $\beta>0$ and $\ho{F}= (\mathcal{F}_t)_{t\in [T]}$ be a filtration.
Let $\{y_{t\alpha}\}_{t\in [T],\alpha\in {\cal A}}$ be real numbers, where $\cal A$ is a finite set.
Let $(\delta_{t\alpha})$ be an $\ho{F}$-predictable process and $(Y_{t\alpha})$ be an $\ho{F}$-adapted process. 
Suppose for each $t\in [T]$, $Y_{t\alpha}$ is\\
{\rm (i)} unbiased: $\ho{E}[Y_{t\alpha}|{\cal F}_{t-1}] = y_{t\alpha}$ for each $\alpha\in {\cal A}$,\\
{\rm (ii)} negative correlated: for any $S\sse \cal A$ with $|S|\ge >1$, we have \[\ho{E}\lb [\prod_{\alpha\in S}Y_{t\alpha}\middle |\  \mathcal{F}_{t-1}\rb] \le 0,\quad {\rm and}\] \\
{\rm (iii)} reasonably bounded: $0\le \beta Y_{t\alpha}\le 2\delta_{t\alpha}$ a.s. for any $\alpha\in {\cal A}$.\\
Then, for any $\delta\in (0,1)$, we have
\[\ho{P}\lb[\sum_{t\in [T]}\sum_{\alpha\in \cal A} \beta \lb(\frac{Y_{t\alpha}}{1+\delta_{t\alpha}} - y_{t\alpha}\rb) \ge \log \frac 1\delta\rb] \le \delta.\]
\end{proposition}
\proof 
It suffices to show that $M_n := \prod_{t=1}^n \xi_t$ is a super-\mtg\ (indexed by $n$) where \[\xi_t = \exp\lb(\sum_{\alpha\in {\cal A}} \beta \lb(\frac{Y_{t\alpha}}{1+\delta_{t\alpha}} - y_{t\alpha}\rb)\rb).\]
In fact, if this is true, then  by Markov's \ineq, 
\[\ho{P}\lb[\prod_{t=1}^n \xi_t \ge \frac 1\delta\rb]  \le \ho{P}\lb[M_n \ge \frac 1\delta \ho{E}[M_n]\rb] \le \delta,\]
which completes the proof.

Now we show that $(M_n)$ is a super-\mtg.
We first use the Cramer-Chernoff method to prove that $\ho{E}_{t-1}[\xi_t]\le 1$ for any $t\in [T]$. 
We will use the following fact: for any $x>0$, we have \begin{align}\label{eqn:112623}
\exp\lb(\frac x{1+\lam}\rb)\le 1+x\le e^x.
\end{align}
Write $\ho{E}_{t}[\cdot]:= \ho{E}[\cdot |{\cal F}_t]$ for each $t\in [T]$. Then, 
\begin{align*} \ho{E}_{t-1}\lb[ \exp\lb(\sum_{\alpha\in {\cal A}}\frac {\beta Y_{t\alpha}}{1+\delta_{t\alpha}}\rb) \rb] 
&= \ho{E}\lb[\prod_{\alpha\in {\cal A}} \exp\lb(\frac {\beta Y_{t\alpha}}{1+\delta_{t\alpha}} \rb)\rb]\\
&= \ho{E}\lb[\prod_{\alpha\in {\cal A}} \lb(1+\beta Y_{t\alpha}\rb)\rb] & \text{by the first \ineq\ in \cref{eqn:112623}}\\
&\le \ho{E}\lb[1+ \beta \sum_{\alpha\in {\cal A}} Y_{t\alpha}\rb] & \text{negative correlation}\\
& = 1+ \beta \sum_{\alpha\in {\cal A}} y_{t\alpha} & \text{by unbiasedness}\\ 
&\le \exp\lb(\beta \sum_{\alpha\in {\cal A}} y_{t\alpha}\rb) & \text{by the second  \ineq\ in \cref{eqn:112623}}.
\end{align*}
Rearranging, we deduce that
\begin{align}\label{eqn:112323} \ho{E}_{t-1}[\xi_t]\le 1
\end{align}
Thus,  by the tower rule, for any $n\ge 1$ we have \[\ho{E}\lb[\prod_{t=1}^n \xi_t\rb] =  \ho{E}\lb[\ho{E}_{n-1}\lb[\prod_{t=1}^n \xi_t\rb] \rb] =  \ho{E}\lb[\prod_{t=1}^n   \xi_n \cdot \ho{E}_{n-1} [\xi_t]\rb]
\le \ho{E}\lb[\prod_{t=1}^{n-1} \xi_t\rb].\]
By induction, we deduce that $\ho{E}\lb[\prod_{t=1}^n \xi_t\rb]\le 1$, and hence $(M_n)$ is a super-\mtg. \qed

To proceed, let us apply the above with clusters as the index ``$\alpha$''.
Specifically, for a fixed RRP $\Pi$, let us decompose $\wh Y_t(a)$ as a sum over the clusters by writing
\begin{align*}
\wh Y_t(a)= \frac 1N \sum_{u\in U} \frac {{\bf 1}(X_{uta}^r =1)}{Q_{uta}^r+\beta} Y_{ut}(Z_t) = \frac 1N  \sum_{C\in \Pi} Y_{Cta}
\end{align*}
where \[Y_{Cta} := \sum_{u\in C} \frac {{\bf 1}(X_{uta}^r =1)}{Q_{uta}^r+\beta} Y_{ut}(Z_t).\]
However, the negative correlation condition (ii) does not hold in general. 
In fact, consider two neighboring clusters $C,C'\in \Pi$ and units $u\in C$ and $u'\in C'$ with $B(u,r)\cap B(u',r)\neq \emptyset$.
Then, for any $a\in [k]$, the exposure mappings $X_{uta}^r$ and $X_{u'ta}^r$ are {\em positively} correlated and therefore $Y_{Cta}$ and $Y_{C'ta}$ are dependent.

We avoid this obstacle by coloring the clusters so that $Y_{Cta}$'s with the same color are independent.

\bdefn[Nine-Coloring]
Fix an $(\ell,r)$-RRP $\Pi$ of $[0,\sqrt N]^2$. 
A mapping $\chi:\Pi\rar \{0,1,2\}^2$ is a valid {\em $9$-coloring} if for any $(i,j)$, $(i',j')$, \[\chi(C_{ij}) = \chi(C_{i'j'}) \quad \iff\quad  i\equiv i' {\rm and}\ j\equiv j' \quad ({\rm mod}\ 3).\]
\edefn

 
We show that two clusters of the same color have independent contributions to $\wh Y_t(a)$.

\begin{lemma}[Independence within  Color Class]\label{lem:indep_same_color}
Let $\chi$ be a valid $9$-coloring of $\Pi$, an $(\ell,r)$-RRP.
Then, for any $C,C'\in \Pi$ with $\chi(c)=\chi(c')$, and $u\in C, u'\in C'$, we have $Y_{Cta}\perp Y_{C'ta}$ for all $a\in [k]$.
\end{lemma}
To see this, note that each $Y_{Cta}$ is determined by the arm assigned to the clusters. 
Formally, conditional on the history up to the $(t-1)^{\rm st}$ round, $Y_{Cta}$ only depends on $\{Z_{C',t}\mid C'\in \Gamma(C)\}$ where $\Gamma(C) = \{C' \cap B(u,r) \neq \emptyset  \text{ for some } u\in C'\}$.
\cref{lem:indep_same_color} then follows by noting that $Y_{Cta}$ and $Y_{C'ta}$ are determined by two {\em entirely} different sets of \rv s, i.e., $\Gamma(C)' \cap \Gamma(C) = \emptyset$.

We use \cref{lem:indep_same_color} and Proposition \ref{prop:cramer_chernoff} to bound the deviation of $\wh R_a$ from $R_a$. 
Recall that the number $m$ of clusters in an $(\ell,r)$-RRP for $[0,\sqrt N]^2$ satisfies $m\ell^2=N$.

\begin{proposition}
[Bounding $\wh R_a -R_a$]
\label{lem:2nd_term} 
It holds that 
\begin{align*} \max_{a\in[k]}\lb\{\wh R_a - R_a \rb\} \le \frac 8{\beta m}\log \frac 1 \delta + \frac T {\beta\ell^2} \psi(r)\quad \text{and}\quad \sum_{a\in [k]} \lb(\wh R_a - R_a\rb) \le \frac 8{\beta m}\log \frac 1 \delta + \frac T {\beta \ell^2} \psi(r).\end{align*}
\end{proposition}
\proof Denote by $N_c$ the (random) number of units in each cluster $c$, then $N_c \le 2\ell^2$ since $r\le \ell/2$. 
Then, 
\begin{align}\label{eqn:011623}
m\lb(\wh R_a - R_a\rb) &= \sum_{t=1}^T \sum_{C\in \Pi} \frac 1{\ell^2} \sum_{u\in C} \lb(\frac {{\bf 1}(X_{uta}^r=1) Y_{ut}(A_t)}{Q_{uta}^r+\beta} - Y_{ut}(a\cdot {\bf 1})\rb)\notag\\
& \le \sum_{t=1}^T \sum_{C\in \Pi} \frac 1{\ell^2} \sum_{u\in C} \lb(\frac {{\bf 1}(X_{uta}^r=1)\cdot (Y_{ut}(a\cdot {\bf 1})+\psi(r))}{Q_{uta}^r+\beta} - Y_{ut}(a\cdot {\bf 1})\rb)\notag\\
&=\sum_{t=1}^T \sum_{C\in \Pi} \frac 1{\ell^2} \sum_{u\in C} \lb(\frac {{\bf 1}(X_{uta}^r=1) Y_{ut}(a\cdot {\bf 1})}{Q_{uta}^r+\beta} - Y_{ut}(a\cdot {\bf 1})\rb)+ \sum_{t=1}^T \sum_{C\in \Pi} \frac 1{\ell^2} \sum_{C\in \Pi} \frac {\psi(r)}{Q_{uta}^r+\beta} \notag\\
&\le \sum_{t=1}^T \sum_{C\in \Pi} \frac 2{N_c}\sum_{u\in C} \lb(\frac {{\bf 1}(X_{uta}^r=1)}{1+\frac \beta{Q_{uta}^r}}Y_{ut}(a\cdot {\bf 1}) - Y_{ut}(a\cdot {\bf 1})\rb) + \frac {mT}{\beta \ell^2}\psi(r),
\end{align}
where the first equality follows since $m\ell^2 = N$, and the final \ineq\ is because $Q_{uta}^r \ge 0$ and $N_c \le 2\ell^2$ a.s.
To apply \cref{prop:cramer_chernoff},
consider an unbiased estimate
\[\widetilde Y_{ut}(a) := \frac {{\bf 1}(X_{uta}^r=1)}{Q_{uta}^r} Y_{ut}(a\cdot {\bf 1}),\] for $Y_{ut}(a\cdot {\bf 1})$. Then, 
\begin{align*}
\eqref{eqn:011623} &= 2\sum_{t=1}^T \sum_{C\in \Pi} \frac 1{N_c} \sum_{u\in C} \lb(\frac 1{1+\frac \beta{Q_{uta}^r}} \widetilde Y_{ut}(a) - Y_{ut}(a\cdot {\bf 1})\rb) + \frac {mT}{\beta\ell^2} \psi(r)\\
&\le 2\sum_{\kappa\in [4]} \sum_{t=1}^T \sum_{C: \chi(C)= \kappa} \lb( \frac 1{1+\frac \beta{P_{ta}}} \frac 1{\ell^2} \lb(\sum_{u\in C} \widetilde Y_{ut}(a)\rb)- \bar Y_{Ct}(a\cdot {\bf 1}) \rb) + \frac {mT}{\beta \ell^2} \psi(r),
\end{align*}
where the \ineq\ is because $Q_{uta}^r \le P_{ta}$.

We conclude by applying the Cramer-Chernoff \ineq\ to the above. 
In \cref{prop:cramer_chernoff}, take 
\[Y_{t\alpha}:= \frac 1{N_c} \sum_{u\in C} \widetilde Y_{ut}(a) \quad \text{and}\quad \delta_{t\alpha}: = \frac \beta {P_{ta}}.\]
Then, condition (i) in \cref{prop:cramer_chernoff} is \satd\ since \[\ho{E}[Y_{t\alpha}] = \ho{E} \lb[\frac 1{N_c} \sum_{u\in C} \widetilde Y_{ut}(a)\rb] = \bar Y_{ct}(a\cdot {\bf 1}).\]
Moreover, for each color $\kappa$, each term in the summation $\sum_{c:\chi(c) = \kappa}$ are \indep, so (ii) is \satd. 
Finally, note that for each $t,a$ we have
\[\beta Y_{t\alpha} = \beta \frac 1{N_c} \sum_{u\in C} \widetilde Y_{ut}(a) = \beta \frac 1{N_c} \sum_{u\in C}\frac {{\bf 1}(X_{uta}^r=1)}{Q_{uta}^r} Y_{ut}(a\cdot {\bf 1})\le \frac {8\beta} {P_{ta}},\]
where the last \ineq\ follows since $Q_{uta}^r\ge P_{ta}/8$ and $0\le Y_{ut}(\cdot )\le 1$. 
Therefore, by \cref{prop:cramer_chernoff}, we conclude that 
\begin{align*}
\beta m\lb(\wh R_a - R_a\rb) &\le 2\beta \sum_{\kappa\in [4]} \sum_{t=1}^T \sum_{c: \chi(c)= \kappa} \lb( \frac 1{1+\frac \beta{P_{ta}}} \frac 1{N_c} \lb(\sum_{u\in C} \widetilde Y_{ut}(a)\rb)- \bar Y_{ct}(a\cdot {\bf 1}) \rb) + \frac {mT}{\ell^2} \psi(r)\\
& \le 8\log \frac 1 \delta + \frac{mT}{\ell^2} \psi(r),
\end{align*}
i.e., 
\[\wh R_a - R_a \le \frac 8{\beta m}\log \frac 1 \delta + \frac T {\beta\ell^2} \psi(r).\]
The proof of (2) is identical by replacing every ``$\sum_{u,t}$'' with ``$\sum_{u,t,a}$''. 
\qed

\subsection{Proof of \cref{thm:main}}
We begin by recalling the regret (w.r.t. a fixed arm $a^*$) decomposition:
\[R - R_{a^*} = (R-\wh R) + (\wh R - \wh R_{a^*}) + (\wh R_{a^*} - R_{a^*}).\]
Let us bound each term separately using the lemmas we have shown so far. 
By \cref{lem:1st_term},
\begin{align}\label{eqn:012324_a}
R - \wh R &\ls \frac {rT}\ell + \beta\sum_{a\in [k]} \wh R_a\notag\\
&=\frac {rT}\ell + \beta \sum_{a\in [k]} R_a + \beta \sum_{a\in [k]} (\wh R_a - R_a) \notag\\
&\le \frac {rT}\ell + \beta kT + \beta \sum_{a\in [k]} (\wh R_a - R_a).
\end{align}
By \cref{lem:EXP3_style,lem:3rd_term}, w.p. $1-\delta$ we have 
\begin{align}\label{eqn:012324_b} \wh R - \wh R_{a^*} &\le \frac {\log k}\eta + \eta \sum_{t,a} P_{ta} \wh Y_t(a)^2\notag\\
&\ls \frac {\log k}\eta + \eta T \lb(1 + \frac km \log \frac 1\delta+ \frac k\beta \psi(r)^2\rb).
\end{align}
Combining \cref{eqn:012324_a,eqn:012324_b},
\begin{align}\label{eqn:112423}
R-R_{a^*} &= (R - \wh R)+(\wh R - \wh R_{a^*})+ (\wh R_{a^*} - R_{a^*})\notag\\
&\ls \lb(\frac {rT}\ell + \beta kT + \beta \sum_{a\in [k]} (\wh R_a - R_a)\rb) + \lb(\frac {\log k}\eta + \eta T\lb(1 + \frac km \log \frac 1\delta + \frac k\beta \psi(r)^2\rb)\rb) + (\wh R_{a^*} - R_{a^*})\notag\\
&\ls \beta kT + \frac {rT}\ell + \frac {\log k}\eta + \eta T\lb(1 + \frac km \log \frac 1\delta + \frac k\beta \psi(r)^2\rb) + \frac 1{\beta m} \log \frac 1 \delta + \frac T {\beta \ell^2} \psi(r),
\end{align}
where the last \ineq\ follows from \cref{lem:2nd_term} and  that $\beta \le 1$.
The statement follows by rearranging the terms into three categories: (i) those that involve $\log \frac 1\delta$, (ii) those that do not involve $\log \frac 1\delta$ but involve $\eta$, and (iii) others.\qed

\section{Experiments}

We consider a $2$-armed setting with $N$ units lying on a $\sqrt N \times \sqrt  N$ lattice. 
We generate unit-level interference as follows. 
Each unit $u\in [N]$ is assigned a random reward $R_{ut}$.
Let $\rho_{ut}$ be the proportion of the five immediate neighbors (counting $u$ itself) assigned arm $1$ at time $t$.
Then, the reward at $u$ is $(2\rho_{ut}-1) c_{ut}$.

In two sets of experiments, we assume that $N=T^2$ and $N=T^3$ \resp, and let $T$ range from $10,20,\dots,50$. 
For each fixed $N,T$, we randomly generate $100$ instances. 
To necessitate exploration, we add large-scale non-stationarity by randomly generating drifts. Each drift  is an $8$-piecewise constant function, where the value on each piece is \indep ly drawn from $U(0,1)$.
To align with the theoretical analysis, we partition the lattice into square-shaped clusters of side length $N^{-1/4}$.
For simplicity, we perform a simplified version of the clustering without randomly assigning the boundary units to nearby clusters.

We compare the switchback version of EXP3-IX (denoted SB) and our clustered randomization-based policy, dubbed EXP3-IX-HT, and denoted ``CR'' in the figures.
We evaluated the performance of the two policies by running each of them $200$ times for each instance.
We then compute the mean and the 95 percentile of the regret for each instance, and average these numbers over the $100$ random instances.

\begin{figure}[h]
\centering
\begin{minipage}{.45\linewidth}
\centering
\includegraphics[width=\linewidth]{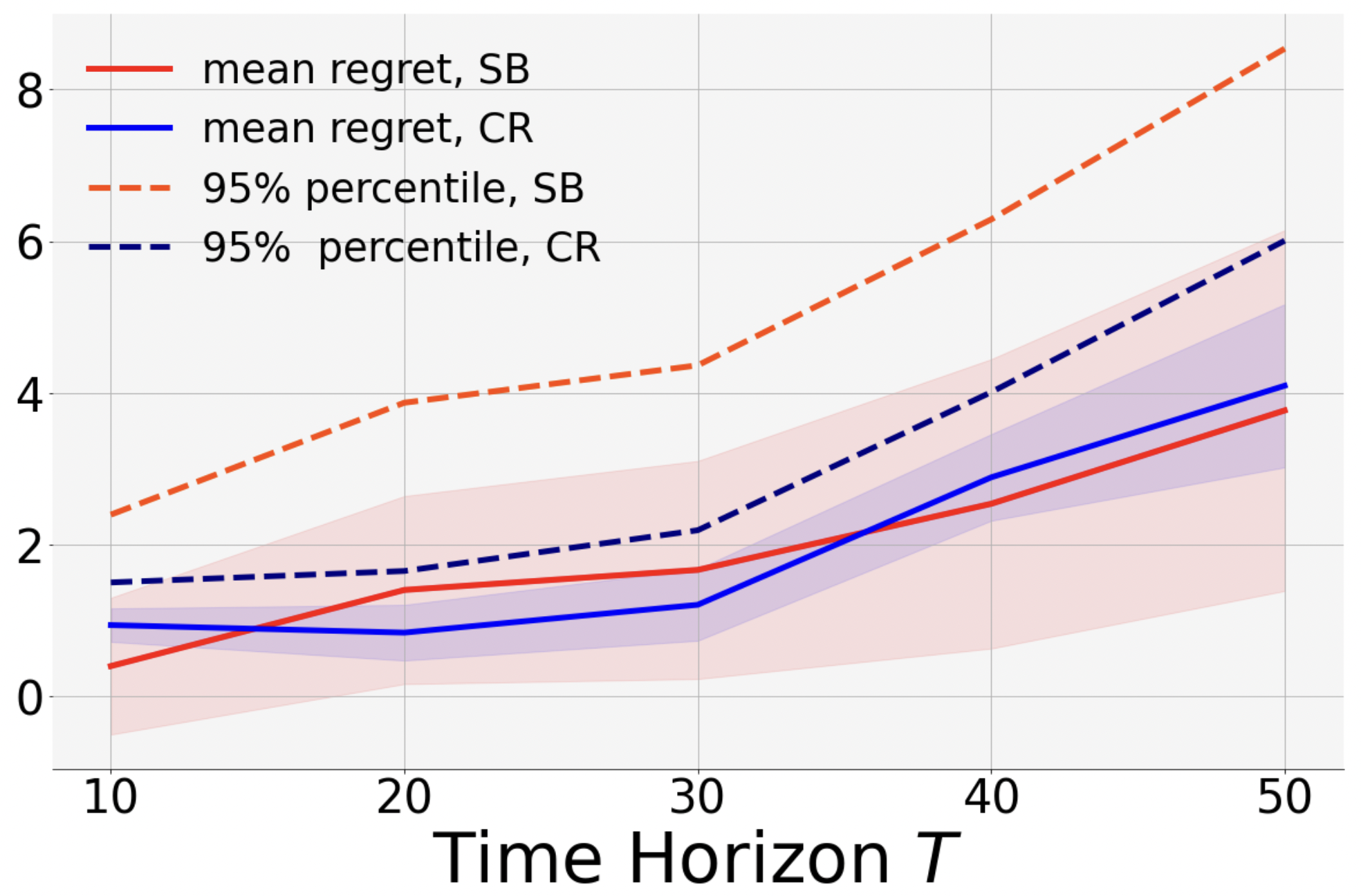}
  \captionof{figure}{$N=T^2$ case}
  \label{fig:N=T^2}
\end{minipage}
\hspace{0.25cm}
\begin{minipage}{.45\linewidth}
  \centering
  \includegraphics[width=\linewidth]{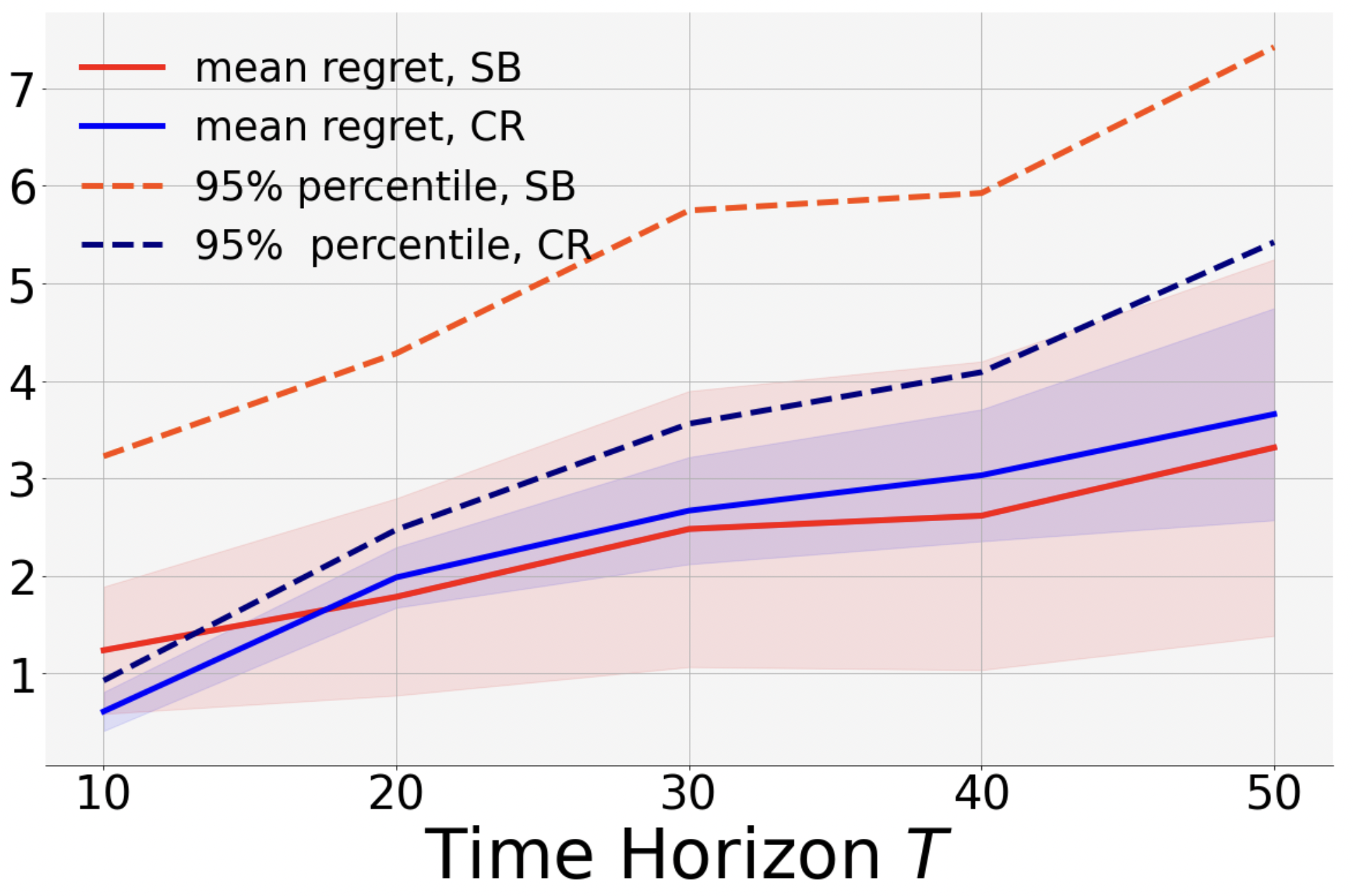}
  \captionof{figure}{$N=T^3$ case}
  \label{fig:N=T^3}
\end{minipage}
\end{figure}

We visualize the results in \cref{fig:N=T^2,fig:N=T^3}.
Consistent with the theoretical analysis, CR outperforms SB in terms of 95\% percentile regret, without sacrificing mean regret. 
Moreover, we observe that when $N=T^3,$ the regret exhibits a smaller deviation.
Finally, we observe that when $N=T^3$, the 95 regret percentile of CR is lower compared to the $N=T^2$ case. 
This is reasonable since a larger $N$ helps reduce the variance in the reward estimation.

\bibliography{causal_ref}
\bibliographystyle{plainnat}
\newpage

\appendix
\section{Proof of Proposition \ref{prop:robustness}}
Consider two cases. \\
{\bf Case 1.} \Sps\ $B(u,r)$ intersects some interior, say $I_{ij}$. Then, $B(r,r)$ intersects at most $2$ strips and $1$ quad. The probability that these 3 regions are assigned to $C_{ij}$ is \[\frac 12 \times \frac 12 \times \frac 14 = \frac 1{16}.\]
{\bf Case 2.} \Sps\ $B(u,r)$ intersects no interior or strips. Then, $B(u,r) \sse Q$ for some quad $Q$ and hence $\ho{P}[B(u,r)\sse C[u]] = 1.$\\
{\bf Case 3.} \Sps\ $B(u,r)$ intersects no interior and exactly $1$ strip $S$. Then, it must also intersect some quad $Q$. Denote by $I_{\alpha_1},I_{\alpha_2}$ the two interiors that $S$ neighbors where $\alpha_1, \alpha_2\in \{1,\dots, B/\ell\}^2$. Then, 
\begin{align*}
\ho{P}\lb[B(u,r)\sse C[u]\rb] &= \sum_{i=1,2} \ho{P}[S\sse C_{\alpha_i}] \cdot \ho{P}\lb[Q\sse C_{\alpha_i} \mid S\sse C_{\alpha_i}\rb] \\
& =\frac 12 \cdot \frac 14 + \frac 12 \cdot \frac 14 = \frac 14.
\end{align*}
{\bf Case 4.} \Sps\ $B(u,r)$ does not intersect any interior, and intersects $2$ strips, denoted $S,S'$, and $1$ quad, denoted $Q$.
Then, $S,S'$ are neighboring the same interior, say $C_{ij}$. Then, \[\ho{P}[B(u,r)\sse C_{ij}] = \ho{P}[S\sse C_{ij}] \cdot  \ho{P}[S'\sse C_{ij}] \cdot  \ho{P}[Q\sse C_{ij}] = \frac 1{16}.\]
The statement follows by combining the above four cases.\qed
\end{document}